\documentclass[manuscript]{IEEEtran}
\IEEEoverridecommandlockouts

\usepackage{amsmath}
\usepackage[ruled,vlined]{algorithm2e}

\usepackage{listings}

\usepackage{tabularx}
\usepackage{tablefootnote}

\newcolumntype{L}[1]{>{\raggedright\arraybackslash}p{#1}}
\newcolumntype{C}[1]{>{\centering\arraybackslash}p{#1}}
\newcolumntype{R}[1]{>{\raggedleft\arraybackslash}p{#1}}

\usepackage{amsmath,amssymb,amsfonts}
\usepackage{algorithmic}
\usepackage{graphicx,caption}
\usepackage{subfig}
\usepackage{textcomp}
\usepackage{xcolor}
\PassOptionsToPackage{hyphens}{url}
\usepackage{hyperref}
\usepackage{url}
\usepackage[super]{nth}
\usepackage{listings}
\usepackage{dsfont}
\usepackage{multirow}
\usepackage[ruled,vlined]{algorithm2e}
\usepackage{tabularx}
\usepackage{tablefootnote}

\usepackage{cite}
\def\BibTeX{{\rm B\kern-.05em{\sc i\kern-.025em b}\kern-.08em
		T\kern-.1667em\lower.7ex\hbox{E}\kern-.125emX}}

\usepackage{amsmath,amsfonts,bm}

\def\eqref#1{equation~\ref{#1}}

\def\1{\bm{1}}

\def\vtheta{{\bm{\theta}}}

\def\vw{{\bm{w}}}

\DeclareMathAlphabet{\mathsfit}{\encodingdefault}{\sfdefault}{m}{sl}
\SetMathAlphabet{\mathsfit}{bold}{\encodingdefault}{\sfdefault}{bx}{n}

\title{Differentiable NAS Framework and Application to Ads CTR Prediction}

\author{\IEEEauthorblockN{Ravi Krishna, Aravind Kalaiah, Bichen Wu, Maxim Naumov, Dheevatsa Mudigere, Misha Smelyanskiy, Kurt Keutzer} \\
	\IEEEauthorblockA{\textit{University of California at Berkeley, Facebook, Inc.}\\
		{\tt\small \{ravi.krishna, keutzer\}@berkeley.edu, \{aravindkalaiah, wbc, mnaumov, dheevatsa, msmelyan\}@fb.com}}
}

\begin{document}

\newcolumntype{L}[1]{>{\raggedright\arraybackslash}p{#1}}
\newcolumntype{C}[1]{>{\centering\arraybackslash}p{#1}}
\newcolumntype{R}[1]{>{\raggedleft\arraybackslash}p{#1}}

\maketitle
\thispagestyle{plain}
\pagestyle{plain}

\begin{abstract}
	Neural architecture search (NAS) methods aim to automatically find the optimal deep neural network (DNN) architecture as measured by a given objective function, typically some combination of task accuracy and inference efficiency. For many areas, such as computer vision and natural language processing, this is a critical, yet still time consuming process. New NAS methods have recently made progress in improving the efficiency of this process. We implement an extensible and modular framework for Differentiable Neural Architecture Search (DNAS)\cite{DNASCode} to help solve this problem. We include an overview of the major components of our codebase and how they interact, as well as a section on implementing extensions to it (including a sample), in order to help users adopt our framework for their applications across different categories of deep learning models. To assess the capabilities of our methodology and implementation, we apply DNAS to the problem of ads click-through rate (CTR) prediction, arguably the highest-value and most worked on AI problem at hyperscalers today. We develop and tailor novel search spaces to a Deep Learning Recommendation Model (DLRM) backbone for CTR prediction, and report state-of-the-art results on the Criteo Kaggle CTR prediction dataset.
\end{abstract}

\maketitle
\thispagestyle{plain}
\pagestyle{plain}

\section{Introduction}
\label{sec:intro}

In recent years, deep learning based approaches have become the \textit{de-facto} standard methods in a number of fields, such as computer vision (CV) \cite{he2016deep}, automated speech recognition (ASR) \cite{he2019streaming}, and natural language processing (NLP) \cite{devlin2019bert}. Building state-of-the-art deep neural networks (DNNs) often involves a significant experimentation process to find the best DNN architecture suited to a task, considering the varying accuracy requirements as well as inference time constraints involved. This has motivated the development of neural architecture search (NAS) techniques, which aim to automate the process of finding the optimal DNN architecture for a given task. NAS has been successfully applied to many DNN application areas, including CV \cite{zoph2016neural, cai2018proxylessnas} and NLP~\cite{chen2020adabert}.

NAS techniques work by searching for the best architecture in a large search space. This is typically judged by comparing their search efficiency to other NAS methods and to random search~\cite{wu2019fbnet}. A common metric used to measure this efficiency is ``GPU-hours'' of search time, which allows us to compare the machine times used by NAS algorithms to arrive at their results.\footnote{Note that this is distinct from the time to potentially train the NAS-determined architecture if the NAS algorithm does not train its candidate architectures during the search process itself.} Many NAS approaches are based on reinforcement learning (RL)~\cite{zoph2016neural, tan2019mnasnet}. They operate in an iterative fashion where an architecture is sampled, evaluated and the learning is fed back into the sampling process. However, RL-based approaches have a higher search cost, as measured by GPU-hours, than gradient-based NAS approaches which aim to perform the NAS process through the gradient-descent optimization of some parameterized computational graph. For example,~\cite{wu2019fbnet}, a gradient-based approach, showed at least $400\times$ speedup over an RL-based approach~\cite{tan2019mnasnet} when optimizing architectures to run image classification on mobile devices.

We apply the DNAS framework designed in this work to the problem of ads click-through rate (CTR) prediction. CTR prediction is the task of estimating the probability that a given user will click on a given ad at a given time. We can then predict estimated CTRs over a subset of our ad inventory and show the ad which maximizes the predicted CTR (a more complex ranking function can also be used~\cite{FBAdsAuctions}). Section \ref{sec:dnas_for_ads:problem} defines the task of CTR prediction formally. Ads CTR prediction, and CTR prediction for arbitrary content more generally, is one of the backbone technologies for the large-scale Internet services, allowing for personalized user experiences. Given that the accuracy of CTR prediction is critical to the revenue of companies offering such services (e.g. a given percentage increase in the AUC of CTR prediction will result in 5x that percentage increase in CTR \cite{rong2020distributed}), there has justifiably been a large investment in both people and machines to improve the models which predict CTR \cite{smelyanskiy2019zion}. At the same time, the large scale of these services, which reach billions of globally distributed users, means that the inference latency and efficiency of CTR prediction models is a non-trivial consideration. Hard SLAs \cite{gupta2020architectural} also mean that the problem naturally becomes one of finding the architecture which maximizes accuracy subject to some latency constraint. A scalable NAS framework would improve the development velocity of of ranking engineers. All of this motivates the application of DNAS to the problem. 

CTR prediction systems have, in the last decade, evolved from systems based on collaborative filtering or large-scale logistic regression \cite{he2014practical} to models incorporating techniques from DNNs such as neural-collaborative filtering (NCF) \cite{he2017neural}, and more recently heavily deep-learning based models. These include the Deep \& Cross Network \cite{wang2017deep}, Deep Interest Evolution Network (DIEN) \cite{zhou2019deep}, Deep Learning Recommendation Model (DLRM) \cite{naumov2019deep}, Time-Based Sequence Model (TBSM) \cite{ishkhanov2020time}, and many others. Most of these systems share a few components, including their input feature types and some common layers, which we outline in more detail in sections \ref{sec:related_work:rec_sys} and \ref{sec:dnas_for_ads:backbone}.

The contributions of this work are as follows. The rest of the paper follows this general structure, with the exception of related work (section \ref{sec:related_work}) and the conclusion (section \ref{sec:conclusion}).
\begin{enumerate}
	\item \textbf{Novel DLRM CTR prediction search spaces} to apply DNAS to CTR prediction. The problem statement and search spaces for our application of DNAS to CTR prediction is contained in section \ref{sec:dnas_for_ads}.
	\item \textbf{Experimental results on a CTR prediction dataset} demonstrating the efficacy of our approach. These results are presented in section \ref{sec:experiments}. Our best result for embedding cardinality search compresses the total size of embedding tables 15.14$\times$ with a relative 0.0012 increase in loss, demonstrating the promise of our approach (see sections \ref{sec:dnas_for_ads:search_spaces:emb_hash_size} and \ref{sec:discussion:emb_card_search}). Our approach discovered recommendation models that beat the state-of-the-art in terms of logloss  with significantly fewer parameters (0.4442 vs.\ 0.447 of~\cite{shi2020compositional}; see \ref{sec:discussion:comparisons}). Moreover, our approach discovered this model using 52$\times$ less computational effort (see section \ref{sec:discussion:comparisons}).
	\item \textbf{An efficient and extensible DNAS framework}, implemented and open-sourced in PyTorch. We implement the DNAS algorithm as it is described in section \ref{sec:dnas_algorithm} and \cite{wu2018mixed, wu2019fbnet} and in such a way that arbitrary supernets for arbitrary application areas can be used. The framework is described in detail in section \ref{sec:framework}, and section \ref{sec:extensions} is dedicated to helping others extend the framework to their application areas.
\end{enumerate}

\section{Related Work \& Background}
\label{sec:related_work}

\subsection{CTR Prediction Systems}
\label{sec:related_work:rec_sys}

CTR prediction systems have historically evolved to leverage the technologies, computational power, and data available at any given time. Early collaborative-filtering based methods \cite{goldberg1992using}, as well as those based on factorization machines \cite{rendle2010factorization}, have evolved into more complex deep learning based methods due to such methods' ability to handle larger and more complex datasets.

Most CTR prediction systems today rely on two kinds of input features: dense and sparse (or categorical). The former are simply real-valued inputs. The latter typically represents a category instance (eg., the item being ranked, an attribute of  interaction). Such categorical features are passed through an embedding table, which maps each category to a vector of real values, with the resulting embedding vectors being passed to the rest of the model. Note that such sparse features can easily have millions or billions of categories.

The metrics used to make decisions concerning the implementations of CTR prediction systems are typically those focused on service revenue and user experience, such as revenue per impression and actual CTR \cite{li2015click}. For our purposes, however, we consider the CTR prediction objective to be logloss, or binary cross-entropy loss, which can be defined for a vector of predictions $p \in (0, 1)^n$ and a vector of binary labels $y \in \{0, 1\}^n$ as:

\begin{equation}
\mathcal{L}(p, y) = \sum_{i = 1}^{n} y_i \log p_i + (1 - y_i) \log (1 - p_i)
\end{equation}

It is important to note that, in contrast to other fields such as computer vision where tradeoffs of many percentage points of accuracy for efficiency gains can be leveraged \cite{dai2020fbnetv3}, recommender systems have the strictest accuracy requirements. A logloss difference of just 0.001 is practically significant~\cite{wang2017deep}. This has influenced the direction of research in the field, which has focused on improving accuracy through new architectural elements designed to leverage certain types of data.

We now examine some of the architectural innovations in this area. The Deep \& Cross Network \cite{wang2017deep} used a parallel deep network (consisting of fully connected or FC layers) and cross network (consisting of novel cross layers) to better model feature interactions. After concatenating the dense features and all embedding vectors into a single vector $x$, we pass this through both networks, concatenate their outputs, and pass that to a fully-connected (FC) layer. A sigmoid function is used to generate the final click probability. The cross layer, as the name suggests, models cross features through use of an outer product: $CrossLayer(x) = xx^{T} w + b + x$ where $x, w, b \in \mathcal{R}^d$ and $w$ and $b$ are weight and bias parameters for the layer respectively.

The Deep Learning Recommendation Model \cite{naumov2019deep}, which will be covered in more detail in section \ref{sec:dnas_for_ads:backbone}, directly leverages pairwise dot products as in a factorization machine \cite{rendle2010factorization}, but then passes these through a multi-layer perceptron (MLP). It uses a ``bottom MLP'' to directly process dense feature inputs, and uses a ``feature interactions layer'' to take pairwise products between embeddings. This is then concatenated with the bottom MLP's output and fed to a top MLP, which outputs a click probability. Note that the DLRM can also use a concatenation-based feature interactions layer.

The Multi-channel user Interest Memory Network (MIMN)~\cite{pi2019practice} focuses specifically on modeling user history information to improve predictive performance. It focuses on designing specific mechanisms to allow for the retention of user history which can then be used to inform the ``user interest representation.'' The authors also focus on the inference time efficiency of their model for user history sequences of up to a length of 150.

\begin{figure*}[htbp]
	\captionsetup{width=.8\linewidth}
	\centerline{\includegraphics[clip, trim=2.0cm 4.0cm 2.0cm 3.0cm, width=2.0\columnwidth]{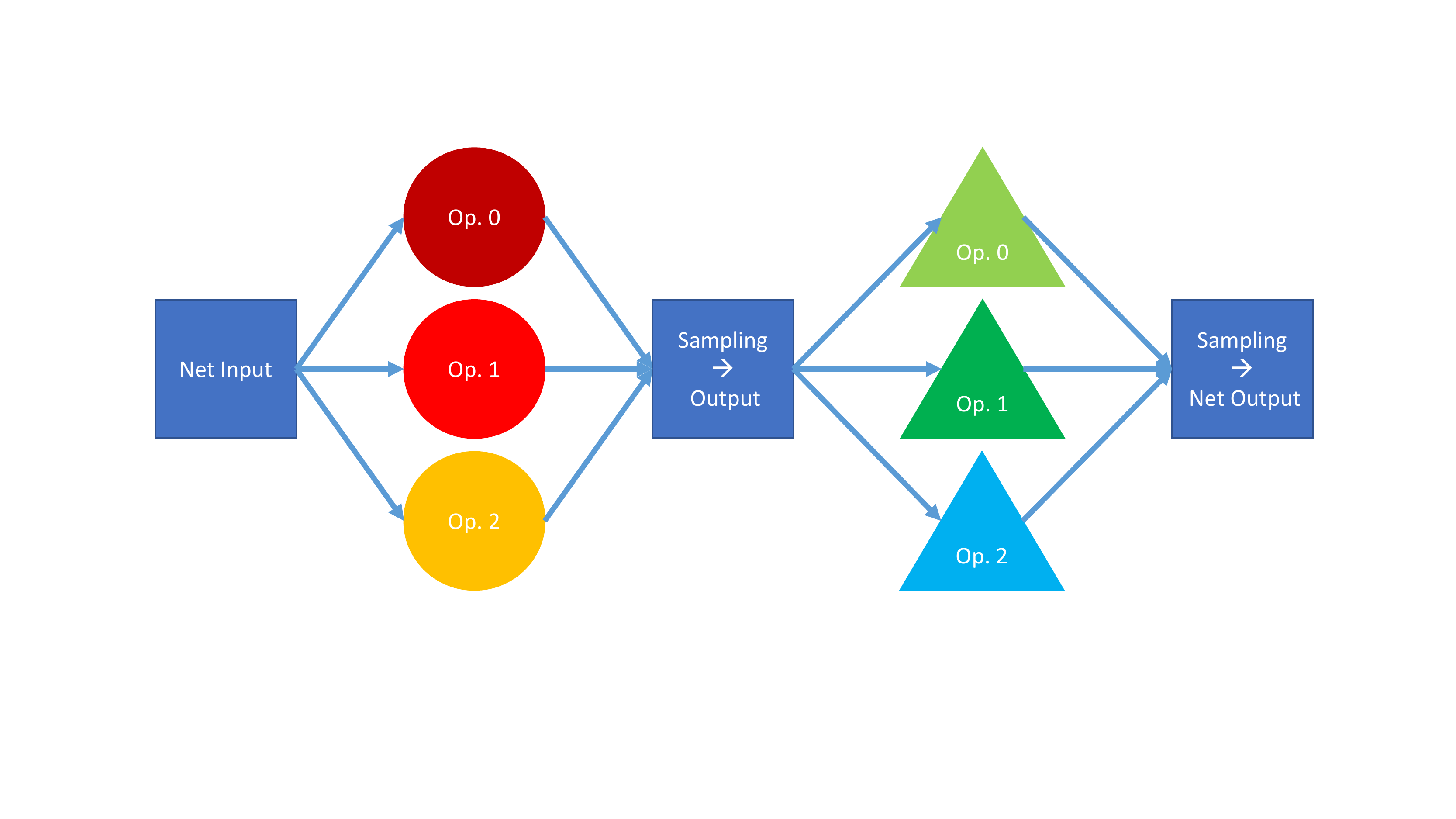}}
	\caption{Representation of a DNAS supernet with two supernet layers and three candidate operators per layer. At the end of each supernet layer, we sample operator weights from the Gumbel Softmax distribution parameterized by the architecture parameters for that layer and take a weighted sum over the operator outputs using those weights, which we send to the next layer.}
	\label{fig:supernet_figure}
\end{figure*}

\subsection{Neural Architecture Search (NAS)}
\label{sec:related_work:nas}

We provide a brief introduction to NAS methods and their application to a few different areas. This is not meant to be a comprehensive summary. Rather, we select methods that we hope will help readers get the most from our work. 

In their work on NAS with reinforcement learning \cite{zoph2016neural}, Zoph et al. propose a controller-trainer reinforcement learning (RL) approach in which a controller samples architectures to be trained and is then updated based the results that these architectures achieve. The authors use a recurrent neural network (RNN) as the controller, which generates the different architectural parameters of the search candidates. As models are trained, the controller is updated via the REINFORCE \cite{williams1992simple} algorithm. One major drawback of this approach is that architectures have to be sampled in a sequential fashion, and that a large number of such candidates must be trained before we can sample high-performing architectures. This is the challenge that gradient-based NAS approaches attempt to remedy.

In contrast to an RL-based methodology, Wu et al.~\cite{wu2018mixed, wu2019fbnet} attempt to complete the entire search process with the training of one supernetwork (covered in much more detail in section~\ref{sec:dnas_algorithm}). Each layer of the supernetwork contains multiple candidate operators with the final output being a weighted average of the individual operators. The  weights are defined as a Gumbel Softmax distribution over a vector of {\em architecture parameters}. The supernet then trains both the operator weights and the architecture parameters. Once this is complete, high-performing architectures can be generated by simply sampling operators from the Gumbel Softmax distribution. As we train the supernet, we anneal the ``temperature'' of the Gumbel Softmax operation such that it becomes closer to hard sampling over time. In this way, the search is completed in one training pass, after which architectures can be generated. ProxylessNAS follows a similar direction of using a supernetwork to essentially conduct the search process~\cite{cai2018proxylessnas}. Both of these approaches allow for directly incorporating latency information in the loss function of the search process by differentiating the expected latency of the network to be sampled from the supernet.

NAS techniques can be applied to many different areas. AdaBERT~\cite{chen2020adabert} focuses on designing architectures specific to individual NLP tasks based on a general pre-trained BERT model. BERT~\cite{devlin2019bert} models are pre-trained on large datasets in a semi-supervised manner and fine-tuned on smaller datasets in a supervised manner; AdaBERT leverages this to design more efficient models during the fine-tuning process. In recommendation systems, AutoCTR~\cite{song2020towards} aims to search over the key ``building blocks'' of a set of DLRM-type recommendation models. This work does not use a supernetwork-based NAS approach, but rather samples architectures. It is able to maintain efficiency by conducting the search process over subsampled versions of the datasets. This approach of using a proxy task is common in NAS~\cite{cai2018proxylessnas}.

\section{DNAS Algorithm}
\label{sec:dnas_algorithm}

In this section, we introduce the DNAS algorithm in detail, in preparation for the following section \ref{sec:framework} which focuses on our implementation of DNAS through our framework. We begin by presenting the structure and usage of the DNAS supernet, the core of the algorithm, then we clarify some points regarding search spaces and sampling, and finally we present the end-to-end DNAS pipeline.

\subsection{DNAS Supernet}
\label{sec:dnas_algorithm:supernet}

The DNAS supernet is what allows the algorithm to arrive at an optimal architecture distribution during a single training process, instead of through sequential sampling and architecture training. An illustration of a supernet is shown in figure \ref{fig:supernet_figure}. Each layer of the supernet consists of a set of \textit{operators}. The input to the layer is fed to each operator, and the outputs of the operators are combined to generate the input to the next layer of the supernet.

Formally, let us denote the $i$th operator at layer $l$ of the supernet as $f_{l,i}$. Then, we can express the output of layer $l$ in terms of its input $x$ as:

\begin{equation}
\label{eqn:weighted_sum}
\sum_{i = 1}^{n} m_{l,i} f_{l,i}(x)
\end{equation}

where the weights $m_{l,i}$ are themselves determined by a Gumbel Softmax over a vector of parameters $\vtheta_l$:

\begin{equation}
m_{l,i} = \text{GumbelSoftmax} (\theta_{l,i} | \vtheta_l)
 = \frac{\exp((\theta_{l,i} + g_{l,i}) / \tau)}{\sum_k \exp((\theta_{l,i} + g_{l,i}) / \tau) }
\end{equation}

The noise $g_{l,i}$ is drawn from a Gumbel(0, 1) distribution. By changing the ``temperature'' parameter $\tau$, we can adjust how ``hard'' or ``soft'' the sampling performed by the Gumbel Softmax function is. As $\tau$ approaches 0, the Gumbel Softmax approaches a hard categorical sampling of each operator, with the probabilities of selection corresponding to those of a regular softmax function over $\vtheta_l$.

During the training of the supernet, we perform soft sampling during each forward pass, allowing us to select a different combination of operators for each sample in the input batch. We exponentially anneal $\tau$ to zero as the supernet training progresses, and we hard sample operators from the distribution at the end of training to result in optimal architectures.

This method of sampling also allows us to incorporate operator latency, FLOP, or other hardware deployment cost information directly into the loss function. The hardware cost of the network is approximated as a sum of the costs for each layer, which is itself calculated using a weighted sum. Thus, the cost of an architecture $a$ which might be soft or hard-sampled from the supernet is:

\begin{equation}
\text{Cost}(a) = \sum_{l} \sum_{i} m_{l,i} \text{Cost}(f_{l,i})
\end{equation}

where $\text{Cost}(f_{l,i})$ is the measured (or calculated) hardware cost for the $i$th operator in supernet layer $l$~\cite{wu2019fbnet}.

The supernet itself is then trained with a loss function which is a combination of the loss for the task on which the supernet is trained as well as the hardware cost loss. In our framework we allow for this to be using either of the below forms:

\begin{equation}
\label{eqn:exponential_loss}
\text{TotalLoss} = \text{TaskLoss} \times \alpha \log (\text{Cost}(a))^{\beta}
\end{equation}

\begin{equation}
\label{eqn:linear_loss}
\text{TotalLoss} = \text{TaskLoss} + \alpha \log (\text{Cost}(a))
\end{equation}

It would be easy for any users of our DNAS framework to implement different $\text{TotalLoss}$ formulations if needed.

\begin{algorithm}
	\SetAlgoLined
	\KwIn{Stochastic super net $G = (V, E)$ with architecture parameters $\vtheta$ and weights $\vw$, 
		searching dataset $\mathcal{X}_\vw$ for weights and $\mathcal{X}_\vtheta$ for architecture parameters, training dataset $\mathcal{X}_{train}$, test dataset $\mathcal{X}_{test}$; initial temperature $T_0$}
	$Q_A \leftarrow \emptyset$ \; 
	\For{$epoch =0, \ldots, N$}{
		$\tau \leftarrow T_0 \exp(-\eta \times epoch)$\;
		Train $G$ with respect to $\vw$ for one epoch\;
		\If{$epoch > N_{warmup}$}{
			Train $G$ with respect to $\vtheta$ for one epoch\;
		}
	} 
	Sample architectures $a \sim P_\vtheta$; Push $a$ to $Q_A$\; 
	\For{$a \in Q_A$}{
		Train $a$ on $\mathcal{X}_{train}$ to convergence\;
		Test $a$ on $\mathcal{X}_{test}$\;
	}
	\KwOut{Trained architectures $Q_A$.}
	\caption{The DNAS pipeline.}
	\label{alg:dnas}
\end{algorithm}

\subsection{Search spaces \& Supernet Training}
\label{sec:dnas_algorithm:search_space_and_supernet_training}

The construction of the supernet is based on the \textit{search space} it is designed to search over. In our case, the search space is the set of operators which can be chosen from at each layer in the network. While we may not search at all over some layers, we could also have different numbers of possible operators at each layer. We use the term \textit{search space group} when we search over a set of search spaces. A search space group might be, for example, searching over embedding dimensions for different features. By contrast, a search space for embedding dimensions should specify precisely which dimensions can be chosen for which categorical features. In section \ref{sec:framework}, we discuss how our framework allows for searching over the search spaces within a search space group.

We also discuss the supernet training procedure before presenting the detailed DNAS pipeline in the following sub-section \ref{sec:dnas_algorithm:pipeline}. The DNAS supernet is actually never trained as a single unit. Instead, we alternatively train 1) the weights of all the different candidate operators $\vw$, and 2) the architecture parameters $\vtheta_l$. Consistent with this methodology, these two sets of parameters are trained on two distinct partitions of the original training dataset. This prevents us from over-fitting the architecture parameters to the training dataset~\cite{wu2019fbnet, wu2018mixed}. It also allows the DNAS algorithm to identify which architectures are prone to over-fitting and reduces the probability that they will be sampled, which is discussed further in section~\ref{sec:discussion:emb_card_search} in the context of our embedding cardinality search space group experiments.

\subsection{DNAS Pipeline}
\label{sec:dnas_algorithm:pipeline}

Our DNAS pipeline illustrated in algorithm~\ref{alg:dnas} is identical to algorithm 3 of \cite{wu2019efficient} except for slight modifications for clarity.

We note a few relevant points from the algorithm. The training dataset $\mathcal{X}_{train}$ is randomly split to generate $\mathcal{X}_\vw$ and $\mathcal{X}_\vtheta$ (see section \ref{sec:framework:training_scripts}). We typically evaluate sampled architectures on a separate validation dataset because many DNAS pipelines may be run for a given dataset, although the test dataset $\mathcal{X}_{test}$ is written in the algorithm. We avoid training the architectural parameters in the early warmup epochs ($N_{warmup}$) so that we do not train the architecture parameters before the weights are trained enough to produce reasonable outputs~\cite{wu2019fbnet}. Otherwise, we might end up with architectures that perform well very early in training but perform poorly by the end of training.

\begin{figure*}[htbp]
	\captionsetup{width=.8\linewidth}
	\centerline{\includegraphics[clip, trim=0.0cm 4.0cm 0.0cm 3.0cm, width=2.0\columnwidth]{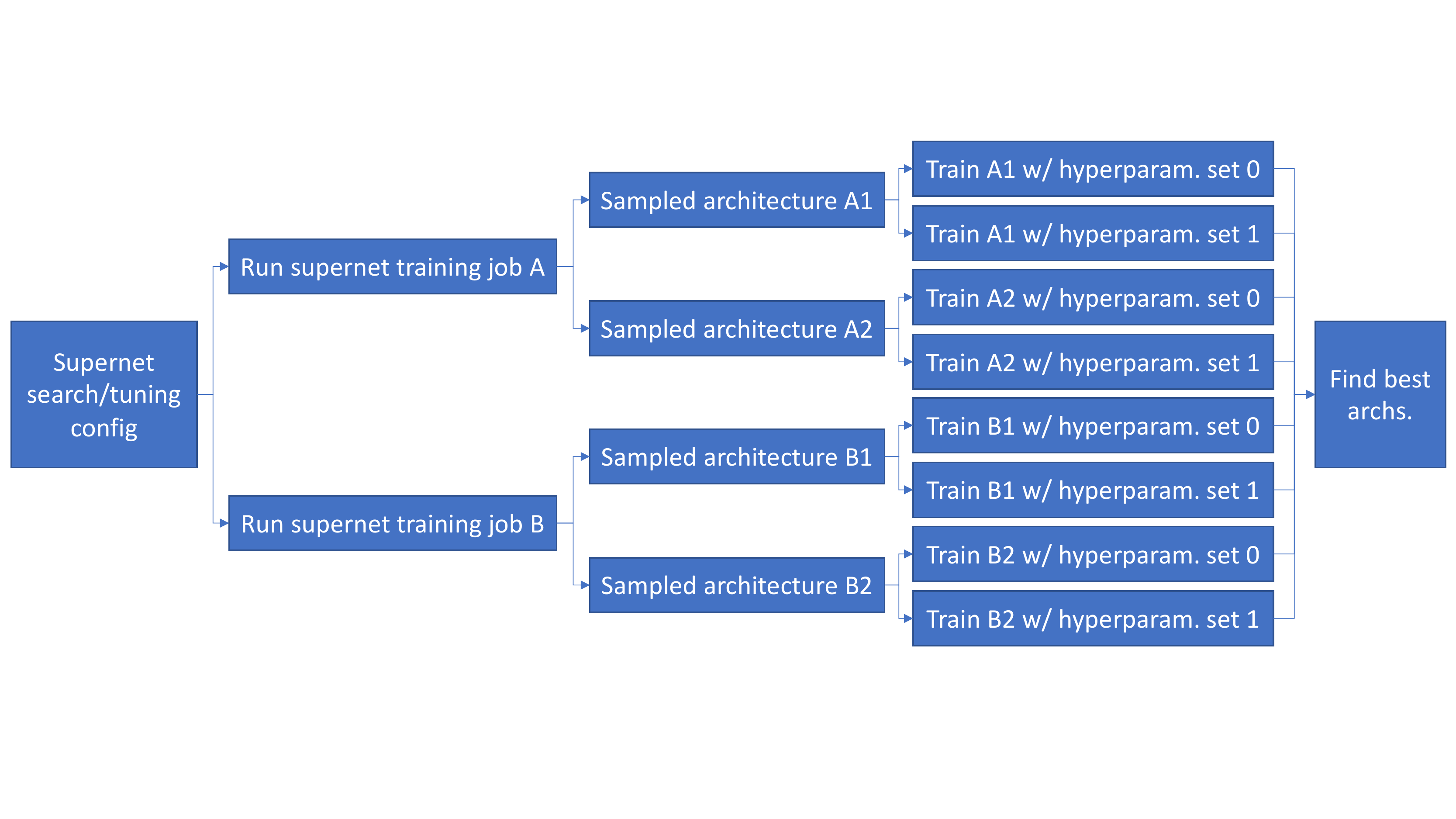}}
	\caption{An illustration of the end-to-end DNAS training/tuning pipeline in our framework. First, separate supernet training jobs are run (for example, exploring different search spaces within a search space group). Then, multiple architectures are sampled from each training run. Finally, each of these architectures is trained with different sets of hyperparameters to derive several sampled architecture training jobs. At the end, we can compare all the results and find the architectures which achieve the best accuracy/efficiency tradeoff.}
	\label{fig:dnas_pipeline_figure}
\end{figure*}

\subsection{DNAS Usage Considerations}
\label{sec:dnas_algorithm:dnas_usage}

In this section we briefly highlight the importance of the various knobs which must be tuned in order for DNAS to achieve good results. These include traditional hyperparameters such as learning rate, weight decay, optimizer choice, etc. It also includes DNAS-specific choices such as the split $\mathcal{X}_\vw$ and $\mathcal{X}_\vtheta$ of the dataset, the structure of the search space group, specific choice of search space, and the parameters to trade off between task loss and hardware latency. While DNAS itself is designed to be a one-pass algorithm in the sense that the search is completed in one single training procedure, different choices of hyperparameters can have a very significant impact on the results achieved. We discuss this further in the context of our ads CTR prediction results, along with other insights regarding developing successful search spaces, in section \ref{sec:insights}.

\section{DNAS Framework Implementation}
\label{sec:framework}

In this section, we provide a detailed look at the implementation of the DNAS algorithm in our framework. The framework is designed to be usable for any application which can be implemented in PyTorch. It is designed to allow for a high degree of tuning and customization, as well as to efficiently run on multi-GPU machines. We structure this section as follows: first, we describe the functionality of the implementation; second, we summarize the major components and how they work together to achieve that functionality; and finally, we present each component of the implementation in detail. We extensively use code snippets from the implementation, which is open sourced and can be found at \cite{DNASCode}.

\subsection{Implementation Functionality}
\label{sec:framework:functionality}

Our framework is designed to allow for large-scale ``push-button'' experimentation and tuning. It natively launches multiple search space training jobs in parallel and multiple sampled architecture training jobs in parallel. That essentially allows the user to specify a search space group and particular search space configurations in addition to the supernet training hyperparameters and sampled architecture training hyperparameters. When the DNAS pipeline completes, the user is able to review all of the results using their own scripts if desired. This is all in pursuit of the goal of allowing both neural net designers and application experts to test new ideas without having to manually tune architectures and hyperparameters.

We illustrate the end-to-end DNAS pipeline as implemented by our framework in figure~\ref{fig:dnas_pipeline_figure}. This shows multiple DNAS supernets being trained in parallel after which multiple architectures are sampled from each one. Subsequently, each of those are trained with different hyperparameters and thus allowing us to find the trained architectures that yield the best tradeoff between accuracy and latency (or other hardware cost parameters, such as the total parameter storage of embedding tables).

\begin{figure*}[htbp]
	\captionsetup{width=.8\linewidth}
	\centerline{\includegraphics[clip, trim=0.5cm 6.0cm 1.0cm 2.0cm, width=2.0\columnwidth]{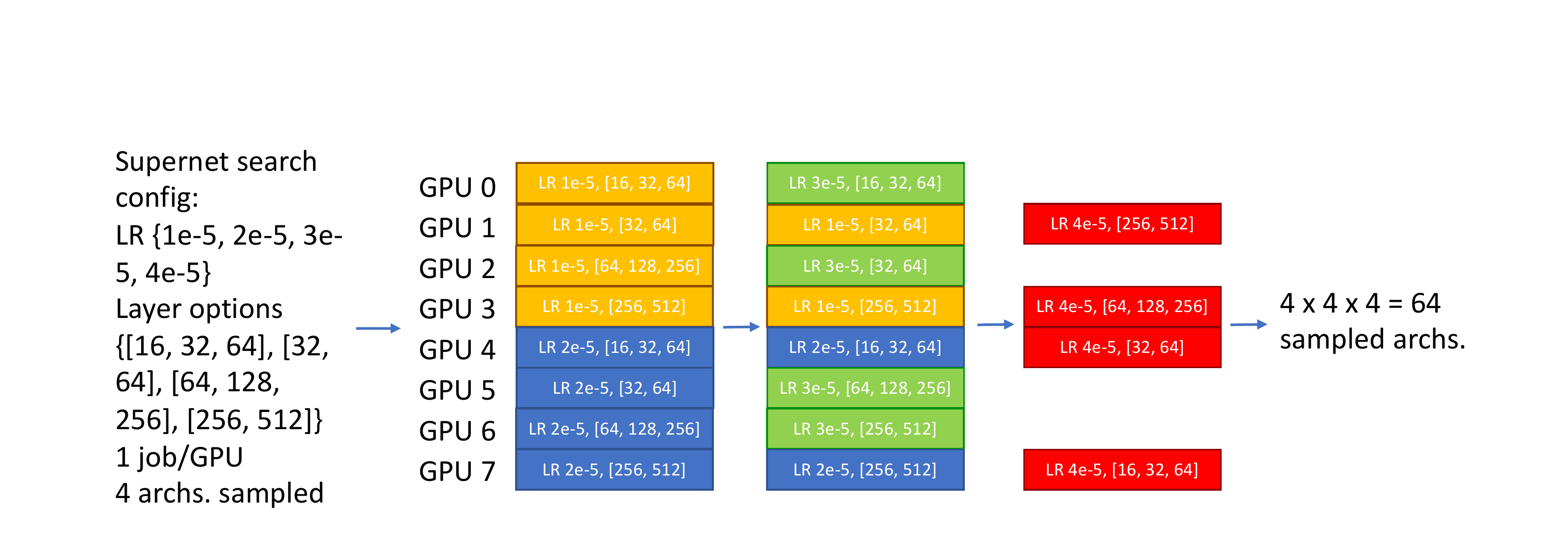}}
	\caption{An example illustrating training of multiple DNAS supernets in parallel to allow for tuning hyperparameters and search spaces. We can simultaneously tune the learning rates and a search space group with different layer sizes to determine best learning rate / search space combination. After training each supernet, we sample four architectures. The training manager, described in section~\ref{sec:framework:training_manager}, immediately launches queued jobs on GPUs as soon as the currently executing job completes, maximizing utilization of compute resources. Different colors for jobs indicate the learning rate used in that job, and help illustrate how the training manager launches jobs asynchronously as the GPUs become available.}
	\label{fig:training_tuning_dnas}
\end{figure*}

We show an illustration of this multi-GPU DNAS training on an 8-GPU machine in figure~\ref{fig:training_tuning_dnas}. This figure illustrates the running of multiple supernet training jobs in parallel to tune the search process. A similar process occurs to train the 8 (in this case) sampled architectures.

\subsection{Implementation Components}
\label{sec:framework:components}

Below we summarize the major components of our implementation, noting the relevant files and classes, and briefly describing each one's functionality. The below sections provide much more detail on each component: the search manager is described in section~\ref{sec:framework:search_manager}, the supernet in section~\ref{sec:framework:supernet}, the training scripts in section~\ref{sec:framework:training_scripts}, training manager in section~\ref{sec:framework:training_manager}, the DNAS pipeline manager in section~\ref{sec:framework:dnas_pipeline_manager} and the utility scripts in section~\ref{sec:framework:utility_scripts}.

The {\it search manager} manages the entire DNAS search process for training a single supernet training from start to finish. This includes initializing the supernet, creating the optimizers for the weights and architecture parameters, alternatively training the weights and architecture parameters, and hard-sampling architectures from the supernet (at any time during training, as specified on input arguments). The search manager is implemented in \texttt{nas\_searchmanager.py} as a single \texttt{SearchManager} class. 

The {\it supernet} is the construct which is trained during DNAS. As mentioned above, the training of the supernet is managed exclusively by the search manager and any code implemented by the user need not have any interaction with the supernet other than initializing it, passing it to the search manager, and reading the architectures sampled from it. We implement a base supernet class, \texttt{SuperNet} in \texttt{nas\_supernet.py}, which provides the Gumbel Softmax sampling and weighted sum utilities needed for any supernet. The user then creates a supernet for their specific search space group (see section \ref{sec:framework:supernet}) as a subclass of \texttt{SuperNet}. In our experiments we used many such supernets, including those implemented in \texttt{nas\_mlp.py}, \texttt{nas\_embedding\_dim.py}, \texttt{dlrm\_supernet.py}, and \texttt{nas\_embedding\_card.py}.

{\it Training scripts} are the front-end interface through which DNAS search jobs and sampled architecture training jobs are launched. The principal purpose of these is to take arguments from the user. For our specific application, we have one \texttt{train\_dnas.py} script and one \texttt{train\_sampled.py} script, whose functions are to take arguments for a single DNAS supernet training and a single sampled architecture training, respectively. The first script results in sampled architectures which are then trained using the second script.

The {\it training manager} handles the parallelization of multiple jobs on a multi-GPU machine and implements the functionality displayed in figure~\ref{fig:training_tuning_dnas}. It is designed to be agnostic to the type of job that it is managing. We can use the same script to manage multiple supernet training jobs and multiple sampled architecture training jobs. We implement the training manager in a single \texttt{tuning.py} script.

Finally, the {\it DNAS pipeline manager} coordinates the entire process. This is mostly a convenience allowing the user to launch a single script and avoid manually launching the training manager first for the DNAS supernet training process and then to train sampled architectures. The pipeline manager takes all arguments related to the supernets, the sampled architectures and data preprocessing. It is implemented in \texttt{run\_dnas\_pipeline.py} script.

In addition to these core components, there will likely be additional {\it utility scripts} needed for an implementation. In our case, we implemented an additional script \texttt{dnas\_data\_utils.py} to perform data preprocessing and splitting operations. Depending on the user's application, they may be able to use existing functions from PyTorch or other libraries and may not need such scripts.

\subsection{Search Manager}
\label{sec:framework:search_manager}

The search manager class is designed with two goals in mind. First, that it should allow the user the maximum possible customizability of the search process, such that the same search manager can be used for any DNAS application with minimal changes. Second, that it should manage the search process end-to-end, requiring minimal user interaction.

To this end, we have implemented the \texttt{SearchManager} class such that there are only two user-facing methods. The first is the constructor, \texttt{SearchManager.\_\_init\_\_()} which allows for the customization. The second is the \texttt{SearchManager.train\_dnas()} method which runs the DNAS process and terminates after all architectures have been sampled. We show an example in listing~\ref{lst:search_manager} of how \texttt{SearchManager} can be used (without complete arguments to \texttt{SearchManager.\_\_init\_\_()} as there are far too many to include in the listing). Instead, we list and categorize these arguments in table~\ref{table:search_manager_args}.

\definecolor{codegreen}{rgb}{0,0.6,0}
\definecolor{codegray}{rgb}{0.5,0.5,0.5}
\definecolor{codepurple}{rgb}{0.58,0,0.82}
\definecolor{backcolour}{rgb}{0.95,0.95,0.92}
\begin{lstlisting}[
float=*,
language=Python,
caption=Example usage of the \texttt{SearchManager} class.,
label={lst:search_manager},
commentstyle=\color{codegreen},
backgroundcolor=\color{backcolour},
keywordstyle=\color{magenta},
numberstyle=\tiny\color{codegray},
stringstyle=\color{codepurple},
tabsize=2, 
numbers=left]
sm_args = {"super_net" : model, ...}
search_manager = SearchManager(**sm_args)
search_manager.train_dnas()
\end{lstlisting}

We split the remainder of this subsection into two pieces: first, we cover the architecture of the \texttt{SearchManager} class and how it works internally; and second, we describe some of the more important arguments to the \texttt{SearchManager} constructor.

\subsubsection{Code architecture}
\label{sec:framework:search_manager:arch}

\texttt{SearchManager} has two main functions: \texttt{train\_dnas()} and \texttt{sample\_archs}. The first manages the entire search process end-to-end and calls the second at the end of each epoch of training in order to sample any architectures that should be at that time. \texttt{train\_dnas} itself relies heavily on \texttt{run\_one\_dnas\_step()} which, as the name indicates, runs one DNAS supernet training step. The main purpose of this function is to simplify the structure of \texttt{train\_dnas()}; it optimizes either the operator weights or the architecture parameters as needed and computes the combination of the task loss and the hardware cost.

The operation of \texttt{train\_dnas()} is determined by several input arguments to \texttt{SearchManager} which are then fed to \texttt{SearchManager.calc\_epoch\_training\_params()}. This function generates a series of configurations for each training epoch, including what should be trained (op. weights or arch. params), the Gumbel Softmax temperature to use during the epoch and how many architectures should be sampled when the epoch is complete. The \texttt{SearchManager} actually allows for fractional epochs so that we can alternate between training the weights and architecture parameters every 0.25 or 0.5 epochs instead of just every epoch. This is useful for our CTR prediction application because the number of epochs used is typically very low (e.g. 5).

\subsubsection{Constructor arguments}
\label{sec:framework:search_manager:args}

While the usage of \texttt{SearchManager} is fairly simple, its usefulness lies in its customizability. To illustrate this, and also to provide some context for users of the code, we list the input arguments to the \texttt{SearchManager} constructor in table \ref{table:search_manager_args}. We organize these by their usage.

\begin{table*}[htbp]
	\caption{Arguments to \texttt{SearchManager}.}
	
	\begin{center}
		\begin{tabular}{L{0.3\columnwidth} L{0.5\columnwidth} L{1.2\columnwidth}}
			\hline
			\textbf{Type} & \textbf{Name} & \textbf{Description}\\
			\hline \hline

			Supernet & \texttt{super\_net} & Supernet PyTorch nn.Module. \\
			 & \texttt{num\_warmup\_} \texttt{epochs} & \# warmup epochs, possible fractional. \\
			 & \texttt{arch\_sampling} & Dictionary mapping keys of \# of arch. params. training epochs completed to \# of archs to sample at that time. \\
			 & \texttt{n\_total\_s\_net\_} \texttt{training\_epochs} & Total \# supernet training epochs. \\
			 & \texttt{n\_alt\_train\_amt} & Amount (possibly fractional) of training after which to switch from training op. weights to arch. params or vice versa. \\
			\hline
			
			Gumbel Softmax params. & \texttt{init\_temp} & Initial temperature. \\
			 & \texttt{temp\_decay} & Temperature exponential decay rate. \\
			\hline
			 
			Misc. & \texttt{host\_device} & PyTorch host device (e.g. GPU). \\
			 & \texttt{update\_lrs\_} \texttt{every\_step} & If true, updates optim. LRs every training step instead of every (posibly fractional) epoch. \\
			 & \texttt{experiment\_id} & Name of search experiment. \\
			 & \texttt{logfile} & Logfile name. \\
			 & \texttt{clip\_grad\_} \texttt{norm\_value} & L2 norm after which gradients are clipped. \\
			\hline
			
			Data & \texttt{w\_dataloader} & PyTorch dataloader for training weights (op. params.). \\
			 & \texttt{m\_dataloader} & PyTorch dataloader for training arch params. (or mask). \\
			\hline
			
			Weights Optimizer & \texttt{w\_optim\_class} & PyTorch optimizer class e.g. \texttt{torch.optim.Adam} for weights training. \\
			 & \texttt{w\_optim\_} \texttt{init\_params} & Dictionary of parameters to initialize weights optimizer. \\
			 & \texttt{w\_optim\_} \texttt{params\_func} & Function which takes supernet as input and returns weights params. to optimize. \\
			 & \texttt{weights\_} \texttt{lr\_lambdas} & A list of (typically lambda) functions, each one corresponding to one parameter group in the weights optimizer, taking the number of weight epochs completed as input and returning a multiplicative factor to be applied to the parameter group's learning rate. \\
			 & \texttt{weights\_} \texttt{initial\_lrs} & List of initial learning rates for weights optim. param. groups. \\
			\hline
			 
			Arch. params. Optimizer & \texttt{m\_optim\_class} & Analogous to corresponding weights argument. \\
			 & \texttt{m\_optim\_} \texttt{init\_params} & Analogous to corresponding weights argument. \\
			 & \texttt{m\_optim\_} \texttt{params\_func} & Analogous to corresponding weights argument. \\
			 & \texttt{mask\_} \texttt{lr\_lambdas} & Analogous to corresponding weights argument. \\
			 & \texttt{mask\_} \texttt{initial\_lrs} & Analogous to corresponding weights argument. \\
		    \hline
		    
			 Loss function & \texttt{loss\_function} & Task loss function assumed to take as input the model predictions and labels. \\
			  & \texttt{use\_hw\_cost} & If false, does not include hardware cost in loss function. \\
			  & \texttt{cost\_exp} & Exponential parameter in exponential hardware cost function as in equation \ref{eqn:exponential_loss}. \\
			  & \texttt{cost\_coef} & Linear parameter in exponential cost (equation \ref{eqn:exponential_loss}) if using exponential hardware cost loss, otherwise linear parameter in linear cost (equation \ref{eqn:linear_loss}). \\
			  & \texttt{exponential\_cost} & If true, uses exponential hardware cost in equation \ref{eqn:exponential_loss}; otherwise, uses linear hardware cost in equation \ref{eqn:linear_loss}. Note both losses take the log of the raw hardware cost before applying a linear or exponential function. \\
			  & \texttt{cost\_multiplier} & Hardware cost is multiplied by this factor before fed to log function. Useful if hardware cost is, for example, measured in seconds and as such is extremely small; we can set the multiplier to 1000 to adjust it upward. \\
			 \hline

		\end{tabular}
	\end{center}
	\label{table:search_manager_args}
\end{table*}

\subsection{Supernet}
\label{sec:framework:supernet}

There are two pieces to our implementation of the DNAS supernet in PyTorch (corresponding to the below two sections). The first is the supernet base class, \texttt{SuperNet}, which provides the basic functionality needed for any other supernet. The second is the supernet for a particular application, for example \texttt{EmbeddingDimSuperNet} in our case to search over embedding dimensions. These specific supernets can also be combined into ``super-supernets'' which can include multiple supernets combined together; in our case, we use this to search over MLP sizes for both the top and bottom MLPs in a DLRM. The same DLRM supernet can also support embedding search spaces, though currently not at the same time as MLP search.

\subsubsection{Supernet base class}
\label{sec:framework:supernet:base_class}

The goal of the base class is to provide the key functionality that any other supernet will need to use. The functions included in \texttt{SuperNet} are thus \texttt{soft\_sample()}, \texttt{hard\_sample()}, and \texttt{calculate\_weighted\_sum()}.

The \texttt{soft\_sample()} function soft-samples from the Gumbel Softmax distribution using some temperature \texttt{temp} creating random architecture weights for each element in the batch as the DNAS algorithm requires. \texttt{hard\_sample()} is called when an architecture needs to be sampled from the supernet and it hard-samples from the Gumbel Softmax distribution to generate a one-hot set of architecture weights for each supernet layer. This operation fixes the operator of the sampled architecture at each layer.

The \texttt{calculate\_weighted\_sum()} function is critical to the operation of any supernet. This function implements the weighted sum of operator outputs from equation \ref{eqn:weighted_sum} which generates the output of each supernet layer. It takes the following as inputs: \texttt{weights}, which should be the output of \texttt{soft\_sample()} for one layer; \texttt{mats}, a list of Pytorch tensors of which the weighted sum should be taken; and \texttt{n\_mats}, which should be equal to \texttt{len(mats)}. Note that every tensor in \texttt{mats} should be of exactly the same size. 

\subsubsection{Specific supernets \& combinations}
\label{sec:framework:supernet:specific_supernet_and_combinations}

When creating a supernet for a specific application, one must first subclass \texttt{SuperNet}, and also set a few class variables in \texttt{\_\_init\_\_}: \texttt{theta\_parameters}, \texttt{mask\_values}, and \texttt{num\_mask\_lists}. We show an example of this with \texttt{EmbeddingDimSuperNet} which searches over the embedding dimensions (see listing~\ref{lst:supernet}). The \texttt{forward()} function of the supernet soft-samples with the specified temperature if the sampling type is soft sampling. We usually choose to store the hardware cost computed at each training iteration as \texttt{curr\_cost}, as this allows for compatibility with ``super-supernets' which expect their sub-modules' \texttt{forward()} functions to return only one tensor. We then later retrieve the \texttt{curr\_cost} variables from each supernet to compute the overall cost.

\begin{lstlisting}[
float=*,
breaklines=true,
postbreak=\mbox{\textcolor{red} {$\hookrightarrow$}\space},
language=Python,
caption=\texttt{EmbeddingDimSuperNet} subclasses \texttt{SuperNet} to search over embedding dimensions. Below is \texttt{EmbeddingDimSuperNet.\_\_init\_\_}.,
label={lst:supernet},
commentstyle=\color{codegreen},
backgroundcolor=\color{backcolour},
keywordstyle=\color{magenta},
numberstyle=\tiny\color{codegray},
stringstyle=\color{codepurple},
tabsize=2, 
numbers=left]
,class EmbeddingDimSuperNet(SuperNet):
  def __init__(self,
               cardinality,
               dim_options):
  """
  Implements an embedding dimension search supernet.

  Adopts an FBNetv2 approach to searching over the number of channels in an embedding by storing a single embedding matrix of the maximum possible dimension and then taking a weighted sum over truncated and then zero-padded versions of the maximum dimension version of an embedding vector.
  """

  # Superclass initialization.
  super(EmbeddingDimSuperNet, self).__init__()

  # Store for later.
  self.cardinality = cardinality
  self.dim_options = dim_options
  self.max_dim = max(dim_options)
  self.num_dim_options = len(self.dim_options)
  self.params_options = nn.Parameter(torch.Tensor([curr_dim * self.cardinality for curr_dim in self.dim_options]), requires_grad=False)

  # Create largest dim embedding matrix.
  self.largest_embedding = EmbeddingDLRM(self.cardinality, self.max_dim)

  # Create other parameters.
  self.theta_parameters = nn.ParameterList([nn.Parameter(torch.Tensor([0.00] * self.num_dim_options), requires_grad=True)])
  self.mask_values = [None] * len(self.theta_parameters)
  self.num_mask_lists = len(self.mask_values)

  # For compatibility with DLRM, store the current cost
  # instead of returning it after each call to forward().
  self.curr_cost = None
\end{lstlisting}

\subsection{Training scripts}
\label{sec:framework:training_scripts}

There are two Python training scripts that we use in our implementation. The first one, \texttt{train\_dnas.py}, takes in arguments via \texttt{argparse} and launches a single DNAS supernet training job by creating a \texttt{SearchManager} and calling \texttt{train\_dnas()}. The second one, \texttt{train\_sampled.py}, also similarly takes in arguments, but launches a single sampled architecture training job on its own (although it could use external objects or libraries if needed). Both scripts need to save a specific file when they have finished running, which may for example contain various information about the training job (e.g. training loss, validation loss, sampled architecture structure, sampled architecture latency). The training manager looks for this file and, if it is found for a particular job, concludes that the job has completed in which case it launches the next job in its queue.

These two scripts are the jobs that the training manager allocates to GPUs. Typically the user would never need to directly launch a \texttt{train\_dnas.py} or \texttt{train\_sampled.py} job. Instead, they provide two configuration files to the DNAS pipeline manager defining what jobs should be launched to tune the search space, supernet hyperparameters and the sampled architecture hyperparameters. The pipeline manager launches the training manager for each step of the DNAS pipeline (supernet training and sampled architecture training) with the appropriate configuration file and the training manager will launch the actual jobs.

Below, we briefly cover some of the arugments to \texttt{train\_dnas.py} and \texttt{train\_sampled.py} as well as their structure.

\subsubsection{DNAS supernet training script}
\label{sec:framework:training_scripts:supernet}

The DNAS supernet training script needs to take in all the arguments necessary to launch a single supernet training job. These include arguments which specify the search space, the hyperparameters (learning rate, weight decay, etc.) for supernet training, data preprocessing methodology, and so forth. Note the distinction mentioned in section \ref{sec:dnas_algorithm:search_space_and_supernet_training} between a \textit{search space} and a \textit{search space group}. A different search space group would require a separate training script, whereas many search spaces can be specifed with arguments to a single script.

\subsubsection{Sampled arch. training script}
\label{sec:framework:training_script:sampled_arch}

The sampled architecture training script needs to take in all the arguments necessary to launch a single sampled architecture training job. These include arguments which specify the hyperparameter (learning rate, weight decay, etc.) for the sampled architecture training, data preprocessing methodology (which may be materially different than the preprocessing methodology for the supernet training) and of course the architecture to be trained itself.

\subsection{Training manager}
\label{sec:framework:training_manager}

The training manager is in a script called \texttt{tuning.py}. It handles launching the jobs specified in a configuration file so as to keep all its assigned GPUs occupied. The inputs to the script are 1) the path to the configuration file itself and 2) an experiment ID which is used in all files related to the jobs that will be launched. It ensures that no files from prior experiments are overwritten. We provide an example configuration file in listing~\ref{lst:tuning_config}.

The first line of listing~\ref{lst:tuning_config} just tells the training manager what script to run and the version of python to use (e.g. users could specify their own virtual environment of python with different packages). Note that this codebase is designed to work with and has been tested only with Python 3. The second line of the configuration file consists of the arguments to the script on the first line. Three parameters must be replaced with special tokens: \texttt{EXPERIMENT\_ID}, \texttt{HOST\_GPU\_ID}, and \texttt{SAVE\_METRICS\_PARAM}. This is because these parameters are replaced with their actual values on-the-fly. For the experiment ID, this is provided when the DNAS pipeline is launched; for the latter two these values change as each job is launched on a different GPU and saves its metrics to a different path.

\begin{lstlisting}[
float=*,
language=Bash,
caption={Example configuration file which can be input to \texttt{tuning.py} which will launch many \texttt{train\_dnas.py} jobs; note that lines 2-9 would be on the same line of the file but are split here for readability. Also note that, for brevity, many arguments are skipped.},
label={lst:tuning_config},
commentstyle=\color{codegreen},
backgroundcolor=\color{backcolour},
keywordstyle=\color{magenta},
numberstyle=\tiny\color{codegray},
stringstyle=\color{codepurple},
tabsize=2, 
numbers=left]
python train_dnas.py
--search_space=emb_card
--embedding_dimension 32
--cardinality_options 1.0 0.1 0.01 0.001
--experiment_id=SAVE_METRICS_PARAM
--host_gpu_id=GPU_ID_PARAM
--weights_lr={0.25, 0.5, 1.0, 1.5, 2.0}
--save_metrics_param=SAVE_METRICS_PARAM
--memory_map
GPU_IDs:0,1,2,3,4
NUM_JOBS_PER_GPU:1
EPOCH_EVAL_FREQ:5,5
\end{lstlisting}

Finally, the parameters which we want to tune are specified by curly brackets as \texttt{weights\_lr}. The values separated by commas inside these brackets represent the different jobs that will be launched. Note that the training manager does a direct string replace of the curly brackets expression with each entry of it. For example, if we want to tune over adding an option, that can be specified inside this option as \texttt{ \{--option\_to\_add, \} }.

The next two lines are reasonable self-explanatory: they outline which GPUs the training manager owns and how many jobs per GPU to launch. The last line is needed for scripts that save output in each epoch (eg., \texttt{train\_sampled.py}). It informs the training manager about the files to look for so that it can aggregate results after each epoch.

The training manager is implemented in a fairly straightforward manner. It first generates all possible jobs to be launched from the contents of the curly brackets expressions. It then puts all of these in a queue and starts launching them on the available GPUs. As soon as it detects the saved file from a particular job (indicating its completion), it launches the next job available in its queue. It can detect files indicating that a particular job has encountered a GPU out-of-memory (OOM) error and add the job to the back of the queue and relaunch it. This feature requires modifications to the script being run with tuning manager so that it catches the OOM exception (usually a \texttt{RuntimeError}) and saves the file indicating the error.

\subsection{DNAS Pipeline Manager}
\label{sec:framework:dnas_pipeline_manager}

The pipeline manager, \texttt{run\_dnas\_pipeline.py}, is a fairly simple script. It takes in all arguments for data preprocessing, as well as the training manager configuration files for supernet and sampled architecture training and manages the launching of the training manager (\texttt{tuning.py}) for both kinds of training. It also updates a logfile with the status (e.g. started supernet training, started sampled architecture training, finished sampled architecture training). In our case, we implement a further script called \texttt{run\_kaggle\_jobs.sh} which fills in many of the arguments to the DNAS pipeline manager which are the same across many of our experiments and only takes as arguments the paths to the configuration files, the search space group, and a few others. However, this additional script is not necessary for all implementations (see the sample extension described in section \ref{sec:extensions}).

\subsection{Utility scripts}
\label{sec:framework:utility_scripts}

Utility scripts needed for a DNAS implementation are likely to be for data preprocessing. In our case, \texttt{dnas\_data\_utils.py} contains the function \texttt{get\_dnas\_dataloaders} which, given command line arguments, preprocesses data and returns Pytorch dataloaders for either DNAS supernet training or sampled architecture training. Note that the test dataloader is only used if specified in the arguments to \texttt{train\_sampled.py}. Generally we would only set this \texttt{--check\_test\_set\_performance} flag once when we test the best architecture from sampled architecture tuning (loading it from a checkpoint).

Other users may find that the built-in dataloaders from PyTorch work with few enough modifications that dedicated utility scripts are not needed. We recommend implementing at least basic utility scripts. For example, this could be to handle splitting the data into weights and architecture parameter training.

\section{Extending DNAS}
\label{sec:extensions}

In this section we cover what it takes to extend the DNAS framework that we have open-sourced to different problems, which may be from any domain and using any backbone network, as long as the search space can be represented as a supernet of some type. This includes both a description of the required implementation components which have to be added for a new application, as well as a code sample which we provide to illustrate how this would actually be done.

\subsection{Required Implementation}
\label{sec:extensions:implementation}

There are only a few implementation requirements which are needed to extend DNAS to another application problem. We enumerate and briefly explain these below.

\begin{enumerate}
	\item \textit{Supernet subclass} corresponding to the search space group. This would be the analog of the \texttt{EmbeddingDimSuperNet} example shown in listing \ref{lst:supernet}. Note that our framework allows this to consist of any valid PyTorch operators. It could be MLP-based, as our supernets are, or convolution based, LSTM-based, etc.
	\item \textit{Sampled architecture module} which is the PyTorch nn.Module that will be used to create and train sample architectures. It should take arguments that are generated when an architecture is sampled from the supernet.
	\item \textit{DNAS training script} to take in arguments and pass them to the search manager. As mentioned in section \ref{sec:framework:training_scripts:supernet}, the main purpose of this script is to make running jobs easier via Python's \texttt{argparse} library.
	\item \textit{Sampled architecture training script} to take in arguments and train a sampled architecture based on them. This script will require a training loop for the given application, which may include any advanced components of the architecture training process that are not present during the search process itself (e.g. changing the batch size during the course of training).
	\item \textit{Configuration files}: one to search across different search spaces within a search space group and supernet training hyperparameters. Another file to search across sampled architecture training hyperparameters. Of course, multiple files are likely to be used as the search over the search space group is refined and better hyperparameters are found.
	\item \textit{Utility script(s)} these are needed for any custom dataloaders or dataset implementations which would be used in the DNAS or sampled architecture training scripts. We implemented this in \texttt{dnas\_data\_utils.py} as mentioned in section \ref{sec:framework:utility_scripts}. While adding a data utility script may not be required for all applications, it is recommended to use one if only to keep the codebase organized given the need for separate weight and architecture parameter training datasets.
	\item \textit{Customizations to the \texttt{SearchManager}} for different types of data, number of inputs to the supernet, different loss functions, etc.
\end{enumerate}

In addition to these minimum implementation components, one can make further changes to the framework to customize it for a particular application. One important area for tweaking is in the combination of the task loss and hardware cost into a single loss function, as different combination functions may suit different applications better.

\subsection{Code Sample}
\label{sec:extensions:sample}

We provide a code sample in our repository (see folder \texttt{sample\_extension}) which shows how our framework could be used to perform search over convolutional networks. In this example, we search over a 5-way classification dataset of random $3 \times 50 \times 50$ images. Note the following correspondence between the implementation requirements mentioned in the above section \ref{sec:extensions:implementation} and the custom files in the sample extension:

\begin{enumerate}
	\item Supernet subclass: \texttt{cnn\_supernet.py}.
	\item Sampled CNN module: \texttt{cnn\_sampled.py}
	\item DNAS training script: \texttt{train\_dnas\_cnn.py}.
	\item Sampled arch. training script: \\ \texttt{train\_sampled\_cnn.py}.
	\item Search space config. file: \\ \texttt{config\_cnn\_dnas\_search}; \\ Sampled arch. config. file: \\ \texttt{config\_cnn\_sampled\_search}.
	\item Data utility script: \texttt{dnas\_cnn\_data\_utils.py}.
	\item Customized search manager script: \\ \texttt{nas\_searchmanager.py}
\end{enumerate}

There are other files in the sample extension that are copied from the main DNAS implementation as these are dependencies required for any implementation (e.g. \texttt{nas\_supernet.py}). We also copied \texttt{utils.py}. However, we deleted some functions from it that are unnecessary in the sample extension (e.g. to calculate latency for an MLP search supernet).

\section{DNAS for Ads CTR Prediction}
\label{sec:dnas_for_ads}

In this section we present our methodology for applying the DNAS framework described in prior sections to the problem of ads click-through rate (CTR) prediction. We start by describing the motivation for and constraints of the problem, and then cover our backbone architecture and search spaces.

\begin{figure*}[htbp]
	\captionsetup{width=.8\linewidth}
	\centerline{\includegraphics[clip, trim=0.0cm 1.0cm 15.0cm 0.0cm, width=1.8\columnwidth]{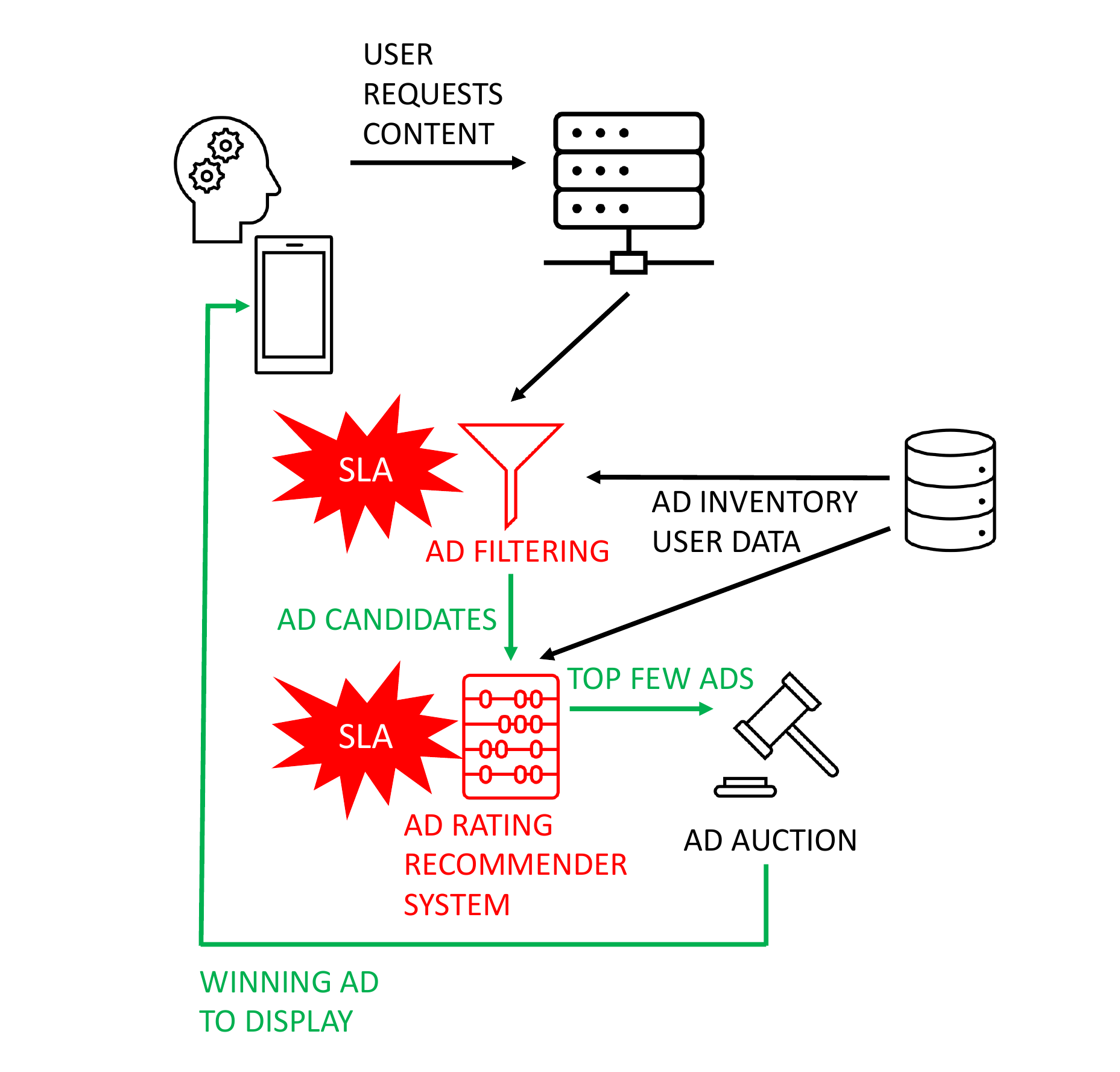}}
	\caption{A typical online advertising retrieval pipeline. The user requests content, for example by loading a page in which an ad needs to be placed. This request is transmitted over some network to a datacenter, where ad inventory and user data (which will be expressed as some dense and sparse features) are combined to first \textit{filter} the ad inventory from thousands or more of candidates to a smaller set using light models, and then rank that set using more complex models \cite{gupta2020architectural}. The top few candidates are then fed to an ad auction which returns the winning ad to be displayed to the user.}
	\label{fig:adproess}
\end{figure*}

\subsection{Problem Statement: Efficient CTR Prediction}
\label{sec:dnas_for_ads:problem}

As shown in figure \ref{fig:adproess}, the process of determining the best ad to show to a user in a given context involves multiple machine learning or deep learning based models, with service level agreements (SLAs) enforced on the services that run these models. This forms the crux of the problem of efficient CTR prediction: recommending the best quality ads while still meeting SLA constraints. These constraints are required to maintain a proper experience both for users of the Internet service as well as for advertisers (i.e. customers) bidding for inventory. SLAs are typically tens to hundreds of milliseconds~\cite{gupta2020architectural}. Some concrete examples of this range can be found in~\cite{gupta2020deeprecsys}.

Given the complexities and constraints of CTR prediction at the scale that is required for major Internet services, it is important to recognize the opportunity for NAS to make material contributions. As noted previously, very small improvements in CTR prediction accuracy can translate to very significant revenue changes for Internet companies, and can equivalently significantly improve user and customer experience. The importance of recommender systems in Internet services, as well as the strict latency constraints they face, motivates the need for a NAS algorithm to make improvements in the pareto-optimal curve of accuracy vs.\ latency.

\subsection{Backbone Architecture}
\label{sec:dnas_for_ads:backbone}

\begin{figure*}[htbp]
	\captionsetup{width=.8\linewidth}
	\centerline{\includegraphics[clip, trim=0.0cm 0.0cm 3.5cm 0.0cm, width=1.8\columnwidth]{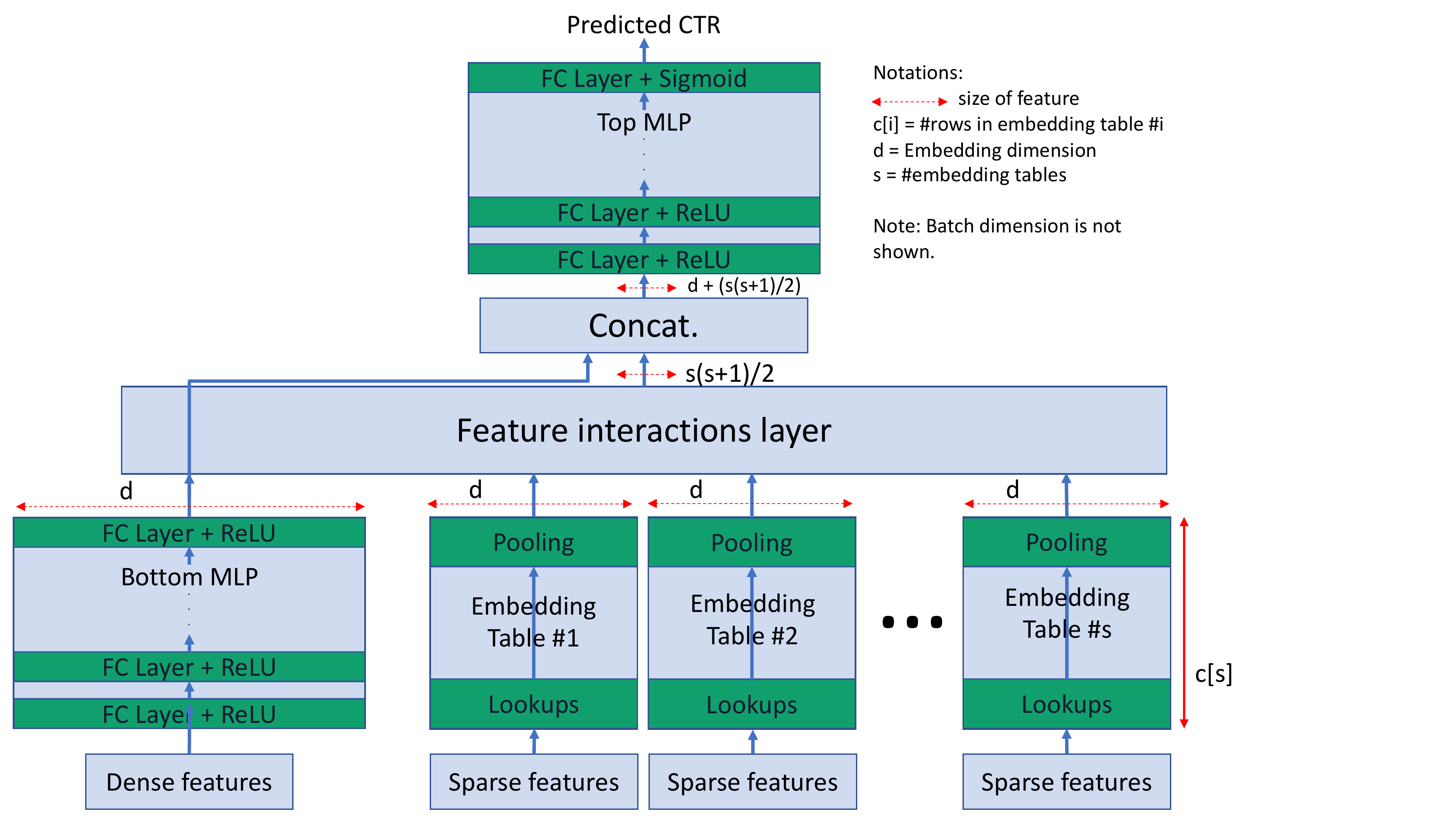}}
	\caption{The DLRM architecture. This figure shows the major components of the DLRM: the embedding tables, the bottom MLP, the feature interactions layer, and the top MLP. The output is the predicted CTR for a particular set of features; these features would incorporate the necessary information about the ad, the user, etc.}
	\label{fig:dlrm}
\end{figure*}

Our backbone architecture, that is the architecture that we use as the starting point for all of our supernets, is the Deep Learning Recommendation Model (DLRM) \cite{naumov2019deep}. We described this model briefly in section \ref{sec:related_work:rec_sys} concerning related work in recommender systems. We will provide more detail in this section. Figure~\ref{fig:dlrm} provides an overview of the model structure.

We  provide further details on the main components of the model using the dimensions noted in figure \ref{fig:dlrm}. Their relevance to the search spaces that will be enumerated in section \ref{sec:dnas_for_ads:search_spaces}:

\begin{enumerate}
	\item \textit{Embedding tables}, which could be numerous, convert a multi-hot sparse feature vector (in our specific application one-hot) to a single dense vector of dimension $d$. Each embedding table has a certain number of rows and may be expressed as a matrix $E \in \mathcal{R}^{c[i] \times d}$. The embedding lookup for a given sparse feature vector $x$ operation is a vector sum $E^T x$. Typically, $x$ will have a small fraction of 1s and almost all 0s, corresponding to certain categories. For example, the location of the 1 may indicate for which user ads are currently being ranked. In practise, due to the sparsity of the matrix-vector product, the operation is implemented as a memory lookup and vector sum.
	\item The \textit{bottom MLP} directly processes the dense feature inputs to the DLRM. These dense features are real-valued and can be represented by $f_d \in \mathcal{R}^{n_d}$, where $n_d$ is the number of dense feature inputs to the model. The bottom MLP consists of sequential fully connected layers and ReLU activation functions.
	\item The \textit{feature interactions layer} combines information from the bottom MLP output and the embedding lookups, all of the same dimension $d$. This layer is designed to be general; however, we use the dot-product based layer in our application. Given $n_s$ embedding tables (sparse features), the unbatched input to the feature interactions layer will be a matrix $F \in \mathcal{R}^{(n_s + 1) \times d}$. The layer output is the flattened (to a vector) upper triangular portion of $F F^T$, possibly including the self-interactions (i.e., diagonal elements).
	\item The \textit{top MLP} takes as input the concatenation of the bottom MLP output and the feature interactions layer output. Similar to the bottom MLP, it consists of sequential FC layers and activation functions. Different from the bottom MLP, the activation function for the last layer is a sigmoid, so that the output of that layer will be a click probability $p_i \in (0, 1)$.
\end{enumerate}

\subsection{Search spaces}
\label{sec:dnas_for_ads:search_spaces}

In this section we discuss the search spaces that we use in this work. We focus on search spaces covering the main components of the DLRM model. We exclude interactions layers from the search space which is considered in more detail in~\cite{song2019autoint, song2020towards}. Our MLP search space is designed to minimize the computational requirement (i.e. FLOPs and latency) needed to achieve a particular accuracy. Further, we believe that our search spaces which focus on embedding tables may be of particular interest for future work, as embedding tables specifically present significant storage and memory bandwidth constraints.

\subsubsection{MLP Search}
\label{sec:dnas_for_ads:search_spaces:mlp}

The particular challenge posed by MLP search is that differing channel dimensions for the outputs of differing operators makes the weighted-sum approach infeasible without changes. One solution is that adopted by FBNetv2 \cite{wan2020fbnetv2} through operator sharing, truncation, and zero-padding. We use a novel solution with a more expansive search space that we term \textit{FC-of-FC}. This solution fully represents every combination of FC operators and thus every possible MLP from the search space directly.

\begin{figure*}[htbp]
	\captionsetup{width=.8\linewidth}
	\centerline{\includegraphics[clip, trim=3.0cm 7.0cm 3.0cm 5.0cm, width=2.0\columnwidth]{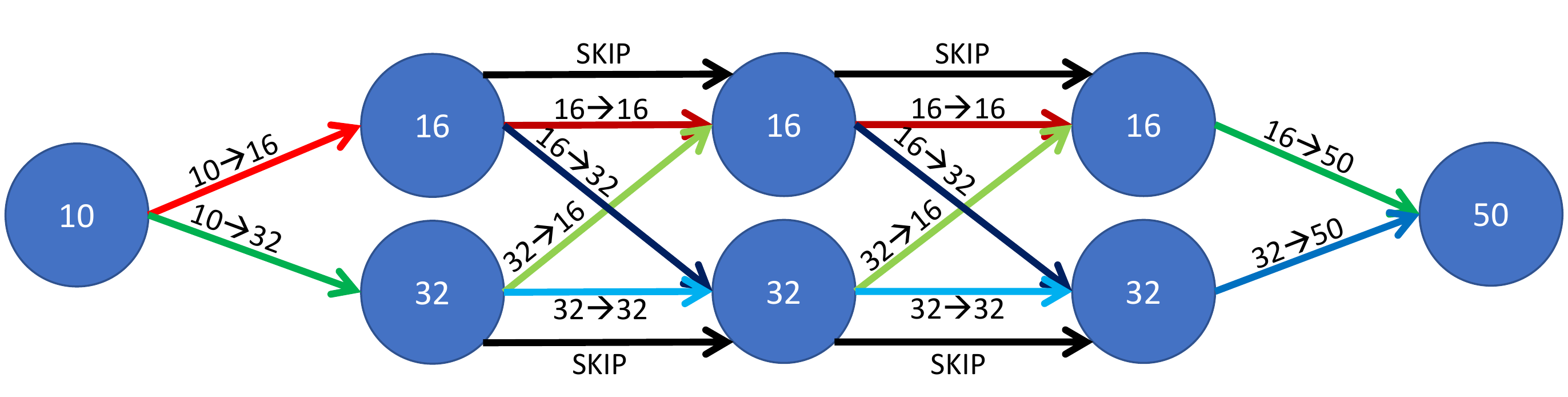}}
	\caption{FC-of-FC supernet for between 2 and 4 FC layers, with potential sizes [16, 32], input and output sizes fixed at 10 and 50 respectively.}
	\label{fig:fc_of_fc}
\end{figure*}

As the name suggests, FC-of-FC consists of a fully-connected network of fully-connected layer operators, allowing for any combination of a fixed number of FC sizes across a fixed number of supernet layers to be sampled. Figure \ref{fig:fc_of_fc} illustrates how an FC-of-FC supernet is constructed using an example.

As shown in figure \ref{fig:fc_of_fc}, every FC-of-FC supernet has a fixed input and output dimension. In between these two endpoints, we search for what sequence of FC sizes (and how many) would be optimal according to some combination of task loss and hardware or inference cost. The figure represents the supernet as a DAG, where nodes are tensors which are weighted sums of operator outputs and operators are represented by edges, which are labeled as input size $\rightarrow$ output size. For each possible output size, we have an operator corresponding to every possible input size. We also have special skip layer operators, denoted in black, which operate between nodes of the same size and skip application of the FC function. Then, the output of a given node is a Gumbel Softmax weighted sum over the outputs of the operators which point to that node.

When we wish to hard-sample networks from this supernet, the process is somewhat more complex than usual. We hard-sample an operator choice from each place where a weighted sum would be conducted, that is, the input to each node. However, this does not determine the network alone. Instead, we work backwards. So, the input to the node labeled 50 might be selected to be 16. Then, we select the edge which was hard-sampled as the input to this 16 node as the prior layer, and proceed until we reach the first either 16 or 32 node, at which point the prior node must be 10 and the edge must be whichever connects 10 to the selected 16 or 32 node.

The latency (or any cost) calculation methodology for this kind of FC-of-FC supernet is also different than would usually be done. We cannot just take a weighted sum because, unlike regular supernets, the operator selections at each layer are not independent. So, we have to multiply the selection probabilities starting from the output of the supernet and then multiply those adjusted probabilities by the latencies of each operator. This is very simple mathematically, but is a bit messy to implement, especially because the Gumbel Softmax operation randomizes selection probabilities within a single training batch. We implement this as a function in \texttt{utils.py}.

In our experiments the MLP supernet is never used on its own, but rather as part of the DLRM. This is done by replacing the fixed bottom and top MLPs with MLP supernets that find the best configurations for these blocks.

\subsubsection{Embedding dimension}
\label{sec:dnas_for_ads:search_spaces:emb_dim}

\begin{figure*}[htbp]
	\captionsetup{width=.8\linewidth}
	\centerline{\includegraphics[clip, trim=2.0cm 5.0cm 2.0cm 3.0cm, width=2.0\columnwidth]{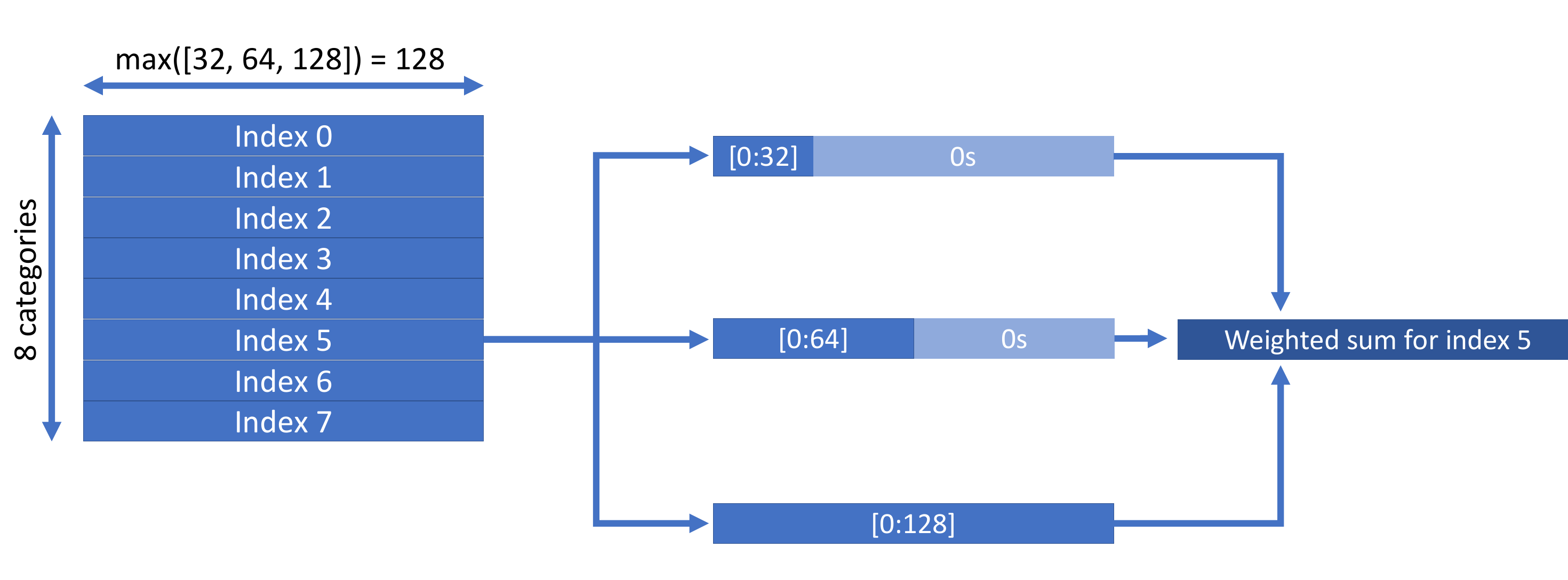}}
	\caption{Illustration of embedding dimension supernet for an embedding table with 8 categories with search over possible dimensions [32, 64, 128]. Here, we show what happens when the embedding at index 5 is looked up from the table. Note that in practice the weighted sum weights are randomized within each batch. Also note that the array slicing above is Python-style e.g. [0:32] denotes all integers $\in [0, 31]$.}
	\label{fig:emb_dim_search}
\end{figure*}

Our embedding dimensions search space is specifically designed to reduce the parameter storage requirements for embedding tables. We do this by searching over a set of possible embedding dimensions for each table. Note that this would be infeasible with the DLRM setup without changes due to the dot-product based feature interactions layer. As shown in figure \ref{fig:emb_dim_search}, we use a zero-padding approach similar to FBNetv2~\cite{wan2020fbnetv2} for embedding dimension search. This works by having a single embedding table of the maximum dimension and replacing elements in an embedding vector beyond each dimension in the search space with zeros. The weighted sum with Gumbel Softmax derived weights is taken over these vectors.

After we have selected the dimensions for a sampled architecture, we maintain the same zero-padding up to the largest dimension of any embedding table in that network. While it would be possible to construct a supernet that required all embedding dimensions to be equal, we believe that setting dimensions specific to certain features allows for more parameter savings. As the embedding dimension search space is a reasonably straightforward application of the DNAS framework, we do not spend much more time on it. We do note, however, that the cost function we use for the embedding dimension search space is the number of parameters in all embedding tables in the network.

\subsubsection{Embedding cardinality}
\label{sec:dnas_for_ads:search_spaces:emb_hash_size}

Our embedding cardinality search space is designed to search over the hash size that is used to reduce the cardinality of sparse features. For example, for a hash size of $H = 100$, we would take $x \mod H$ before looking up a given index $x$, which effectively reduces the number of categories in the feature to at most $H$. This hash size represents a tradeoff between the degree of personalization an embedding table affords and the amount of storage that it requires.

\begin{figure*}[htbp]
	\captionsetup{width=.8\linewidth}
	\centerline{\includegraphics[clip, trim=2.0cm 1.0cm 3.0cm 2.0cm, width=2.0\columnwidth]{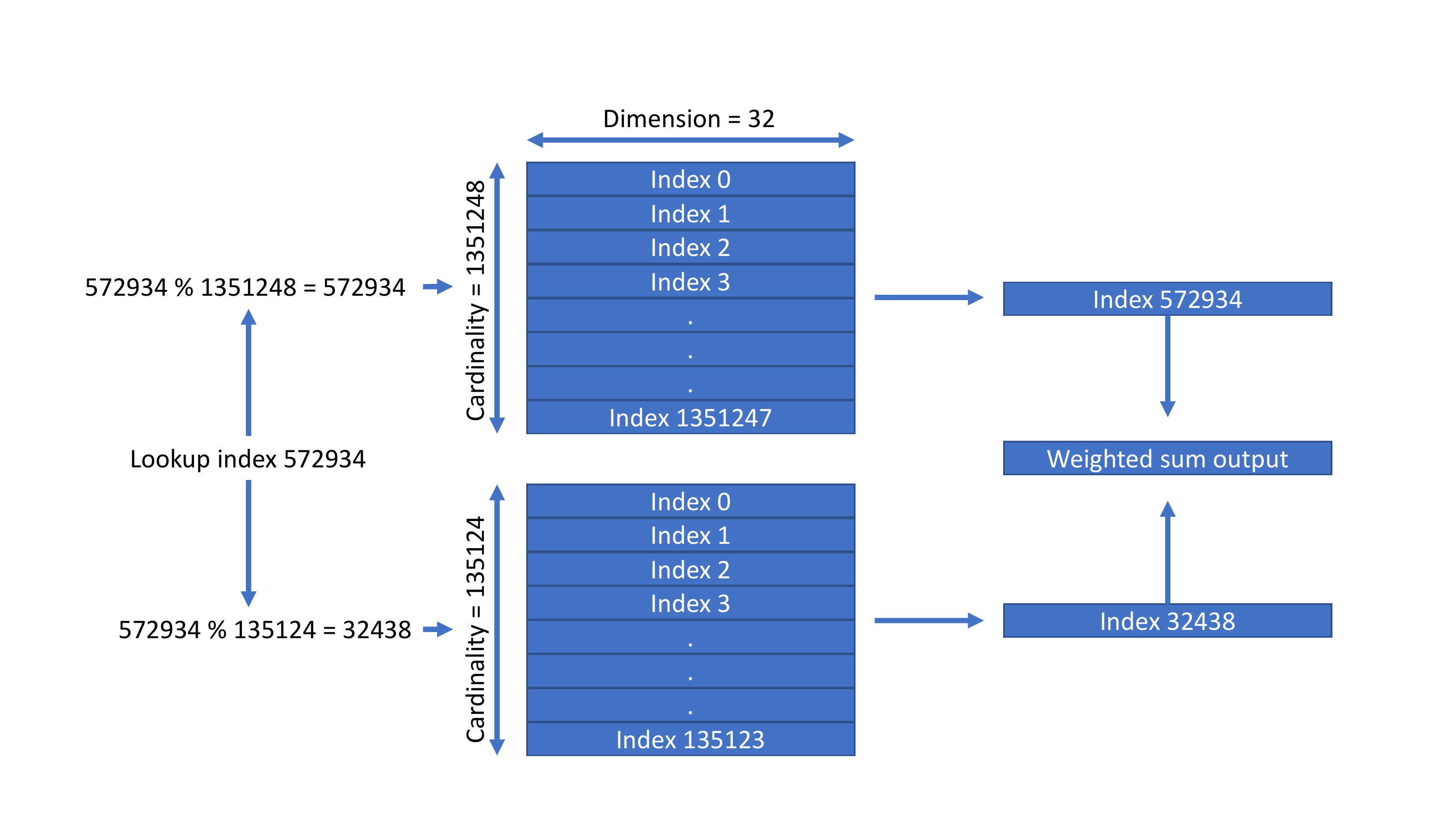}}
	\caption{Illustration of embedding cardinality supernet for an embedding table with dimension 32 with search over possible cardinalities from the original (1351248) to 10x less (135124). Here, we show what happens when the embedding at index 572934 is looked up from the tables and then combined via a weighted sum. The original index is looked up from the original table, and the index 32438 determined via hashing is looked up from the table with a 10x lower hash size. The resulting two embedding vectors are then combined.}
	\label{fig:emb_card_search}
\end{figure*}

The design of the search space is fairly simple. We store separate embedding tables (with the same dimension) for each possible hash size. Then we look up the vectors at index $x \mod H_i$ for each possible hash size $H_i$ in the search space and take a weighted sum over these vectors. Figure~\ref{fig:emb_card_search} provides an illustration of this. While it is possible to combine these different candidate tables into a single table and look up the different indices within that table, we felt that this would not allow for an effective search. Further, the storage overhead at search time is likely to be minimal because in our experiments we typically search over hash sizes decreasing by factors of 10 from the original cardinality of each feature. This means that the total storage requirement is less than or equal to $\frac{1}{1 - 0.1} = 1.11 \times$ that of the original feature.

\section{Experiments}
\label{sec:experiments}

\subsection{Dataset}
\label{sec:experiments:dataset}

In this subsection we introduce the dataset we use in this work, Criteo Kaggle, as well as the preprocessing that we perform on this dataset. 

\subsubsection{Criteo Kaggle}
\label{sec:experiments:datasets:criteo_kaggle}

The Criteo Kaggle dataset\footnote{http://labs.criteo.com/2014/02/kaggle-display-advertising-challenge-dataset/} was originally released by the advertising company Criteo as part of a Kaggle challenge. Because the test set labels from this challenge were never released to the public, the training set is itself split and used in many works as a benchmark for CTR prediction \cite{wang2017deep, naumov2019deep, song2019autoint, song2020towards}. We refer to this henceforth as the dataset.

The dataset itself consists of ~46M chronologically ordered click records from a 7-day period. Each record consists of 13 dense features and 26 sparse or categorical features as well as a binary click label, although some features may be missing from any given record. The categorical features themselves have a wide range of cardinalities. The minimum cardinality is just 3, while the maximum is 10131227, and the total number of categories across all sparse features is 33762577. Depending on the embedding dimension used, the total sparse feature storage can easily be in the gigabytes or even tens of gigabytes (and note that this dataset is orders of magnitude smaller than industry-scale datasets). Approximately 74.4\% of records in the dataset have a 0 (i.e., no-click) label and the remaining ones have a 1 (i.e., click) label.

In accordance with \cite{wang2017deep} and especially \cite{naumov2019deep} as we directly leverage the DLRM Criteo Kaggle data processing code in our \texttt{dnas\_data\_utils.py}, we split the dataset as follows. The first 6/7 are used as training data and the last 1/7 is split randomly (i.e. not chronologically) into 50/50 validation and test dataset. Note that the chronological separation of the training from the validation and test dataset is important because CTR prediction models face the same issue in practice: changing data distributions over time. The validation and test datasets are drawn from the same day so that they are i.i.d.

\subsubsection{Preprocessing}
\label{sec:experiments:datasets:preprocessing}

The preprocessing we perform is reasonably straightforward and contains no novelty. We briefly describe it below for completeness. We follow the preprocessing methodology used in~\cite{naumov2019deep} which in turn closely follows the preprocessing used in~\cite{wang2017deep}.

For dense features, missing features are replaced with -1, and the rest of the features are transformed by the function $f(x) = \ln (1 + x)$.

We do not perform any subsampling of the data, either to increase the proportion of click labels, or to decrease the total number of records used during training time. Doing the latter especially would reduce the search time, and this is used in \cite{song2020towards}. That being said, our \texttt{train\_dnas.py} and \texttt{dnas\_data\_utils.py} script support both subsampling methodologies and the DLRM data processing itself supports the first.

\subsection{Model and Search Configuration}
\label{sec:experiments:config}

\begin{table}[htbp]
	\caption{Backbone search architectures. \textbf{TBS} = \textbf{T}o \textbf{B}e \textbf{S}earched, as in \cite{wu2019fbnet}}
	
	\begin{center}
		\begin{tabular}{L{0.1\columnwidth} L{0.2\columnwidth} L{0.2\columnwidth} L{0.125\columnwidth} L{0.125\columnwidth}}
			\hline
			\textbf{Search space} & \textbf{Bottom MLP} & \textbf{Top MLP} & \textbf{Embedding dimensions} & \textbf{Embedding cardinalities }\\
			\hline \hline
			
			MLP Search & \textbf{TBS} & \textbf{TBS} & All 32 & All 20K \\
			\hline
			Emb. dim. & 13-1024-1024-1024-128 & 479-1024-1024-1024-1 & \textbf{TBS} & All 20K \\
			\hline
			Emb. card. & 13-1024-1024-1024-32 & 383-1024-1024-1024-1 & 32 & \textbf{TBS} \\
			\hline
			
		\end{tabular}
	\end{center}
	\label{table:backbone_archs}
\end{table}

This section presents the architectural parameters that we used for our experiments with all three search space groups mentioned in subsection \ref{sec:dnas_for_ads:search_spaces}. Table~\ref{table:backbone_archs} presents the backbone architectures and the components that are part of the search process for each search space group. We now describe the precise search spaces, as well the tuning process that was used in conjunction with each. However, we do not cover all the parameters here. They may be found directly in the open-sourced code tuning configuration files or as default parameters specified in the code.

\subsubsection{MLP search configuration}
\label{sec:experiments:config:mlp_search}

Both the bottom and top MLP layer size options were [128, 256, 512, 1024]. Both MLPs are limited to 5 layers in our search. This means, for example, that one possible bottom MLP configuration of the maximum length might be [13, 128, 128, 1024, 1024, 32]. Because the MLP search space group incorporates layer skipping, a possible bottom MLP configuration of the minimum length might be [13, 128, 32]. We tuned the temperature decay rate and weights LR arguments in \texttt{train\_dnas.py}. The former was one of [0.1, 0.2] and the latter was sampled from [1.0, 1.5, 2.0]. We tuned only the LR when training sampled architectures. It was allowed to be one of [0.25, 0.5, 1.0, 2.0].

\subsubsection{Embedding dimension search configuration}
\label{sec:experiments:config:emb_dim}

The dimension options for each sparse feature were [8, 16, 32, 64, 128]. We tuned the same hyperparameters in the same manner as we did for the MLP search space, for both supernet training and sampled architecture training.

\subsubsection{Embedding cardinality search configuration}
\label{sec:experiments:config:emb_card}

The cardinality options were [1.0, 0.1, 0.01, 0.001], expressed as a factor of reduction from the original cardinality found in the dataset. This means that for a feature with an original cardinality of $10^4$, we would search over candidate cardinalities [10000, 1000, 100, 10]. We tuned the same hyperparameters in the same manner as we did for the MLP search space for both supernet training and sampled architecture training.

\subsection{Results}
\label{sec:experiments:results}

We now present the results from the experiments described previously in the section. Most of our analysis uses validation logloss as our task performance metric, with various efficiency metrics used corresponding to the search spaces groups tested. We also conduct test-set evaluations on the most accurate (overall) as well as most efficient (selecting the tuning configuraiton which performs best on the validation set) architectures for each search space group. This is a total of 3 search space groups $\times$ 2 architectures each $=$ 6 test set results. Note this is a very small fraction of the total sampled architecture results: in total we have 3 seach space groups $\times$ 2 temperature decay options $\times$ 3 weights LR options $\times$ 4 sampled architectures $\times$ 4 sampled training LR options = 288 architectures, all of which have validation loss calculated at every one of 6 training epochs.

We first present statistics on the validation loss, calibration\footnote{Calibration for predictions $p$ and labels $y$ is defined as $\frac{\sum_{i = 1}^n p_i}{\sum_{j = 1}^n y_i}$. Calibration is considered an indication of a model's quality and keeping it close to 1.0 is critical to a model's commercial potential \cite{he2014practical}.}, and efficiency metrics for the three different search space groups. These are found in tables \ref{table:mlp_search_stats}, \ref{table:emb_dim_stats}, and \ref{table:emb_card_stats} for MLP search, embedding dimension search, and embedding cardinality search respectively. Note that these statistics are over each sampled architecture's minimum validation loss epoch across the 6 epochs of training.

\begin{table}[htbp]
	\caption{Statistics for MLP search.}
	
	\begin{center}
		\begin{tabular}{L{0.1\columnwidth} L{0.1\columnwidth} L{0.2\columnwidth} L{0.2\columnwidth}}
			\hline
			\textbf{Statistic} & \textbf{Validation logloss} & \textbf{Validation calibration dist. from 1.0} & \textbf{Bottom and top MLP FLOPs} \\
			\hline \hline
			
			Min & 0.4467 & 0.0027 & 9.927 $\times 10^5$ \\
			\hline
			Mean & 0.4508 & 0.0143 & 1.879 $\times 10^6$ \\
			\hline
			Median & 0.4514 & 0.0138 & 1.806 $\times 10^6$ \\
			\hline
			Max & 0.4534 & 0.0294 & 3.276 $\times 10^6$ \\
			\hline
			
		\end{tabular}
	\end{center}
	\label{table:mlp_search_stats}
\end{table}

\begin{table}[htbp]
	\caption{Statistics for embedding dimension search. Note that average dimension is a valid metric because all cardinalities are fixed at 20K. Average dimension is directly proportional to total embedding storage.}
	
	\begin{center}
		\begin{tabular}{L{0.1\columnwidth} L{0.1\columnwidth} L{0.2\columnwidth} L{0.2\columnwidth}}
			\hline
			\textbf{Statistic} & \textbf{Validation logloss} & \textbf{Validation calibration dist. from 1.0} & \textbf{Avg. emb. dim.} \\
			\hline \hline
			
			Min & 0.4470 & 0.0016 & 67.07 \\
			\hline
			Mean & 0.4504 & 0.0145 & 79.54 \\
			\hline
			Median & 0.4509 & 0.0145 & 81.08 \\
			\hline
			Max & 0.4532 & 0.0299 & 89.54 \\
			\hline
			
		\end{tabular}
	\end{center}
	\label{table:emb_dim_stats}
\end{table}

\begin{table}[htbp]
	\caption{Statistics for embedding cardinality search.}
	
	\begin{center}
		\begin{tabular}{L{0.1\columnwidth} L{0.1\columnwidth} L{0.2\columnwidth} L{0.2\columnwidth}}
			\hline
			\textbf{Statistic} & \textbf{Validation logloss} & \textbf{Validation calibration dist. from 1.0} & \textbf{Total \# categories} \\
			\hline \hline
			
			Min & 0.4451 & 0.0013 & 1.24 $\times 10^6$ \\
			\hline
			Mean & 1.437 & 0.1717& 1.294 $\times 10^7$ \\
			\hline
			Median & 0.4467 & 0.0206 & 1.304 $\times 10^7$ \\
			\hline
			Max & 20.55 & 2.9 & 2.368 $\times 10^7$ \\
			\hline
			
		\end{tabular}
	\end{center}
	\label{table:emb_card_stats}
\end{table}

\begin{figure*}[htbp]
	\captionsetup{width=.8\linewidth}
	\centerline{\includegraphics[clip, trim=0.0cm 0.0cm 2.0cm 0.0cm, width=2.0\columnwidth]{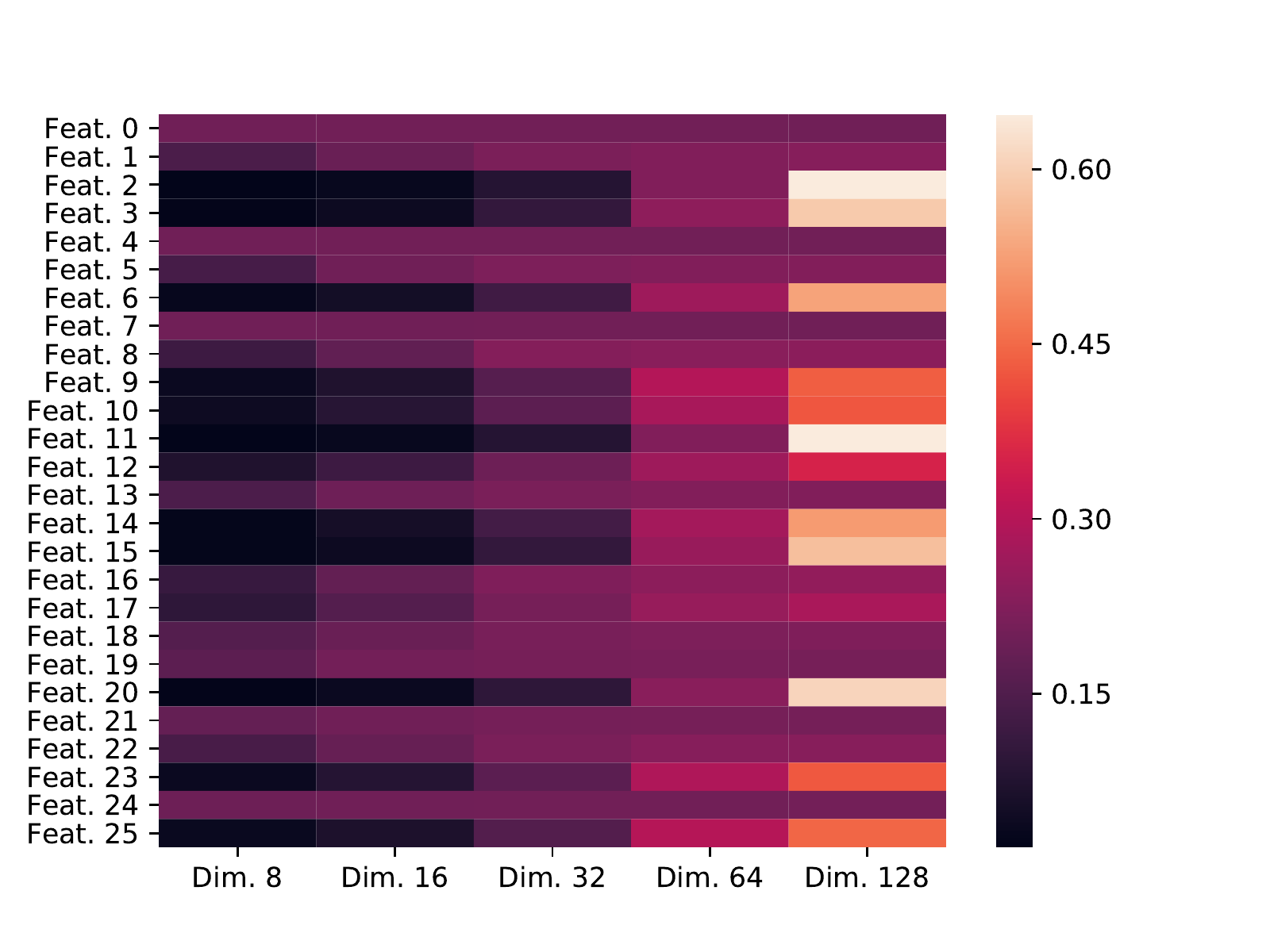}}
	\caption{Heatmap representing the Gumbel Softmax parameters of the embedding dimension supernet which yielded the lowest validation loss architecture. Vertical axis numbers represent index of categorical feature; horizontal axis has length 5 and represents the relative weight of dimensions 8, 16, 32, 64, and 128 respectively. Note that we take a softmax over the vector parameterizing the Gumbel Softmax to normalize the values for plotting, such that the values correspond directly to operator sampling probabilities, as can be seen in the scale to the right. This graph was generated using \texttt{seaborn} and \texttt{matplotlib}.}
	\label{fig:emb_dim_heatmap}
\end{figure*}

\begin{figure*}[htbp]
	\captionsetup{width=.8\linewidth}
	\centerline{\includegraphics[clip, trim=0.0cm 0.0cm 2.0cm 0.0cm, width=2.0\columnwidth]{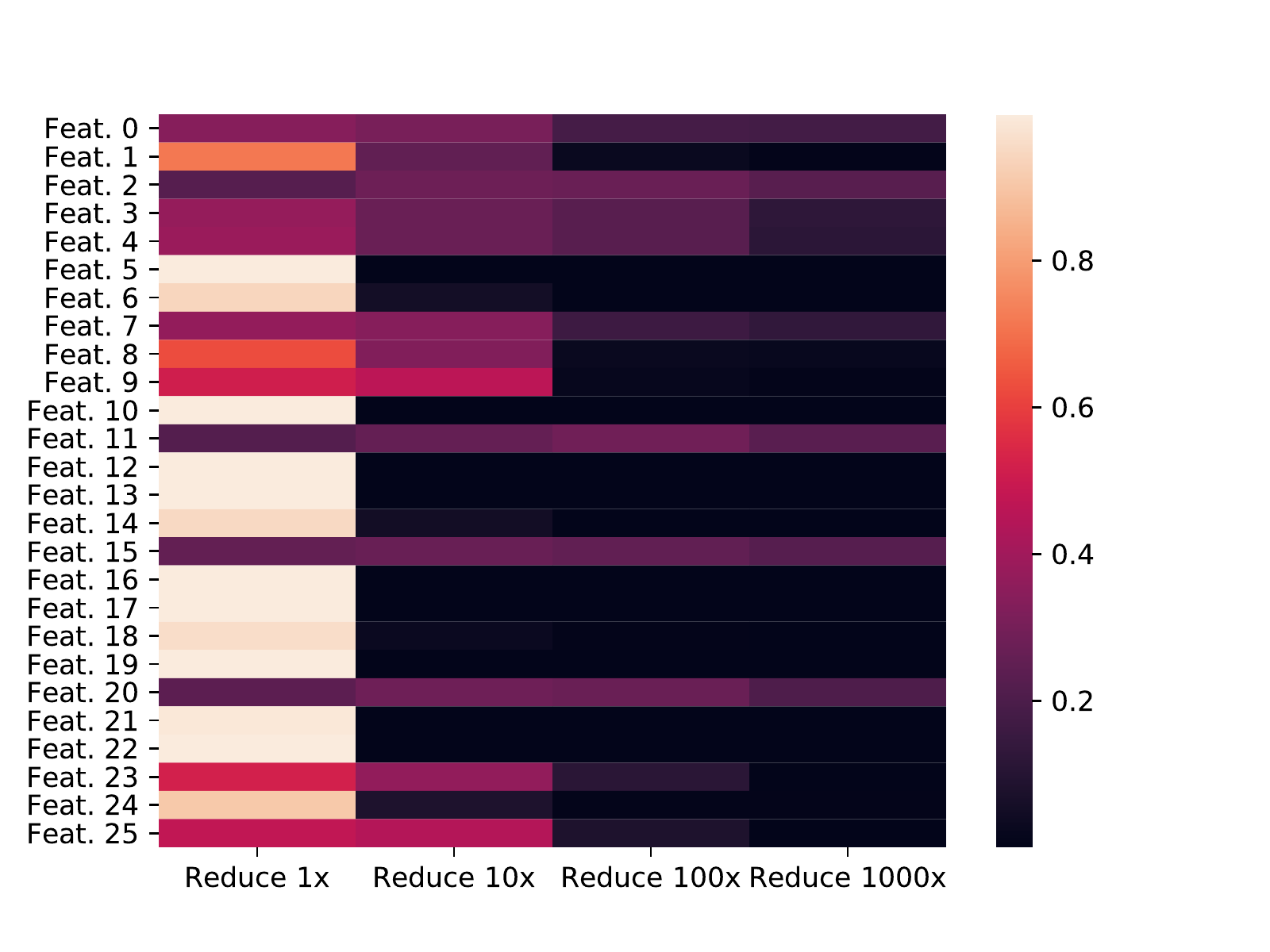}}
	\caption{Heatmap representing the Gumbel Softmax parameters of the embedding cardinality supernet which yielded the lowest validation loss architecture. Vertical axis numbers represent index of categorical feature; horizontal axis has length 4 and represents the relative weight of cardinality reduction factor 1.0, 0.1, 0.01, and 0.001 respectively. Note that we take a softmax over the vector parameterizing the Gumbel Softmax to normalize the values for plotting, such that the values correspond directly to operator sampling probabilities, as can be seen in the scale to the right. This graph was generated using \texttt{seaborn} and \texttt{matplotlib}.}
	\label{fig:emb_card_heatmap}
\end{figure*}

For the purposes of analyzing and attempting to interpret the actual DNAS search process, we also provide heatmaps representing the weights parameterizing the Gumbel Softmax operations in supernets for the best-performing architecture (by loss) in each search space. Please see these in figures \ref{fig:emb_dim_heatmap} and \ref{fig:emb_card_heatmap} for the embedding dimension and cardinality search spaces respectively. Note that these heatmaps do not depict the MLP search space which is discussed more in subsection \ref{sec:discussion:mlp_search}. Finally, we report the test set results mentioned above in table in \ref{table:test_results}. The discussion and analysis of the results is presented in the following section \ref{sec:discussion}.

\begin{table}[htbp]
	\caption{Test set results for selected architectures. Efficiency metric is total MLP FLOPs for MLP search, average dimension for embedding dimension search, and total \# categories for embedding cardinality search. Note that for this table we put the actual calibration instead of its distance from 1, as there is no need to calculate statistics.}
	
	\begin{center}
		\begin{tabular}{L{0.1\columnwidth} L{0.1\columnwidth} L{0.2\columnwidth} L{0.2\columnwidth}}
			\hline
			\textbf{Arch.} & \textbf{Test logloss} & \textbf{Test calibration} & \textbf{Efficiency metric} \\
			\hline \hline
			
			MLP lowest loss & 0.4457 & 0.9906 & 1.12 $\times 10^6$ FLOPs \\
			\hline
			MLP most efficient & 0.4520 & 0.9868& 9.927 $\times 10^5$ FLOPs \\
			\hline
			Emb. dim. lowest loss & 0.4460 & 0.9808 & Avg. dim. = 71.69 \\
			\hline
			Emb. dim most efficient & 0.4463 & 0.9824 & Avg. dim. = 67.08 \\
			\hline
			Emb. card. lowest loss & 0.4442 & 0.9916 & 1.877 $\times 10^7$ categories \\
			\hline
			Emb. card most efficient & 0.4454 & 0.9875 & 1.24 $\times 10^6$ categories \\
			\hline

		\end{tabular}
	\end{center}
	\label{table:test_results}
\end{table}

\section{Discussion of Results}
\label{sec:discussion}

\subsection{MLP search}
\label{sec:discussion:mlp_search}

We first note that we did not provide a heatmap illustration of the architecture parameters for the MLP search space. This is because the architecture parameters for this search space, as is explained in section \ref{sec:dnas_for_ads:search_spaces:mlp}, do not operate across an entire layer but rather weigh the inputs to an operator of a particular size. Thus, these architecture parameters are not directly interpretable. They also do not directly correspond to operator sampling probabilities due to the way in which FC-of-FC works (sampling probabilities are not independent across layers).

From the statistical results in table \ref{table:mlp_search_stats}, we can see that the MLP search process has performed well, achieving a minimum validation log loss of 0.4467. We can also see a ~3.3$\times$ range of FLOPs between the minimum and maximum, indicating that the search space itself would be useful in reducing actual computational requirements. Interestingly, the test set results show much less variation across FLOPs between the most accurate and most efficient model. In fact, the most accurate model uses close to the minimum number of FLOPs found in any of the MLP architectures. The architectures achieve very different logloss with the most accurate achieving 0.4457 and the most efficient achieving 0.4520.

Due to the minimal difference in FLOPs, we believe that this is due to a difference in hyperparameters between the models and not due to any effect of the lower representational capacity of the more efficient model. This may also point to a potential lack of sampling accuracy of FC operators, or insufficient training of the architecture parameters, thus leading to operators being sampled suboptimally.

\subsection{Embedding dimension search}
\label{sec:discussion:emb_dim_search}

We start by examining the results from table \ref{table:emb_dim_stats}. We can see a significant variation in the validation logloss: a range of [0.4470, 0.4532]. The calibration values also follow the logloss numbers, with worse calibration (farther from 1.0) seen for architectures with higher logloss. We can also see a significant range of the average embedding dimension, from 67.07 to 89.54. Also, notably, the mean and median average dimensions are closer to the maximum dimensions than the minimum, suggesting that the optimization of dimensions is non-trivial. Interestingly, the average embedding dimension is \textit{higher} for the architecture that achieves the logloss of 0.4532 than it is for the architecture that achieves the logoss of 0.4470. This suggests that rather than simple compression, what is required to both decrease loss and to decrease model size is compressing the correct features by the correct amount.

On this point, we can look at the architecture parameters represented in figure \ref{fig:emb_dim_heatmap}. DNAS places an emphasis on using larger embedding dimensions for the features at indices 2, 11, 20, and to a lesser extent, 3, 6, 9, 10, 14, 23, and 25. The actual cardinalities of the features are [1460, 583, 10131227, 2202608, 305, 24, 12517, 633, 3, 93145, 5683, 8351593, 3194, 27, 14992, 5461306, 10, 5652, 2173, 4, 7046547, 18, 15, 286181, 105, 142572]. The average log base 10 cardinality of the features with indices listed above is 5.473, and the average for those not is 2.368 (the numbers are 6.925 and 3.124 respectively if we exclude the second set of indices). Needless to say, this difference of ~100x clearly shows that DNAS has identified that high-cardinality features need a larger embedding dimension in order to result in embedding with the necessary representational ability.

Looking at the test set results, we can also see that it is proper optimization of the dimensions for specific features that makes models perform well on the CTR prediction task. The best-performing model achieves a 0.4460 test logloss, but is only ~7\% larger in parameter storage than the most efficient model, which actually achieves a very similar test logloss: 0.4463. Note that both of these logloss numbers are quite low themselves.

\subsection{Embedding cardinality search}
\label{sec:discussion:emb_card_search}

Again, we start by examining the results from table \ref{table:emb_card_stats}. One immediate observation is that the maximum loss is 20.55, which is an absurd value that can only indicate complete learning failure or divergence during training. As a result of such outlier values, the mean logloss is a useless indicator, as is the mean calibration. Also important to note is the degree of variation in the embedding cardinality: we can see a $>20\times$ range in the number of categories, offering the largest opportunities for model compression of all of the search spaces groups used in this work. We also note that the minimum validation logloss is the best we have observed so far: 0.4451.

Now, looking at figure \ref{fig:emb_card_heatmap}, we can see that there are certain features for which DNAS has found that the original cardinality should be maintained. These fall at the indices 5, 6, 10, 12, 13, 14, 16, 17, 18, 19, 21, 22, and also to a lesser extent 24. The features listed have an average log base 10 cardinality of 2.424, and those not listed have an average log base 10 cardinality of 4.702. Clearly, DNAS has realized that it makes sense not to try to reduce the storage used by low-cardinality features, both because this saves little storage space, but also because it is likely to significantly hurt task performance.

We now examine the test set results of this search space. The architecture that achieves the lowest loss achieves a test set logloss of 0.4442, the lowest results achieved in al search spaces. It achieves a calibration of 0.9916, which is also the closest to 1 out of all search spaces. However, it contains ~19M categories, resulting in a large model. What is even more impressive is that the architecture with the \textbf{minimum} number of categories is actually fairly close in logloss, with a test set result of 0.4454, which is the second best of all test set results across all search spaces. This architecture has 15.14$\times$ fewer categories (and thus uses the same factor less embedding table storage) than the architecture with the lowest loss, and yet it increases test loss by just 0.0012. While this would be significant in a commercial context \cite{wang2017deep}, this result shows that DNAS has great promise as a tool to help maximize the performance that can be achieved in a given storage budget for embedding tables.

We also note our belief that part of the value of the embedding cardinality search space is that it allows DNAS to, at some level, remove feature information which contributes primarily to overfitting. Because the weights  and architecture parameters training datasets are separate, DNAS has the ability to recognize features which are not generalizing well and reduce the probability that they are sampled with the same cardinality. Reducing the cardinality will reduce the overfitting, providing both a storage reduction and a loss improvement. We believe that this may also be part of why the architecture with 15.14$\times$ fewer categories than the best-performing one does not perform too much worse; that is, much of the trimmed categories may have been contributing more to overfitting than generalized performance.

\subsection{Comparisons to Prior Work}
\label{sec:discussion:comparisons}

Our test set results of 0.4442 and 0.4454 for the lowest loss and most efficient embedding cardinality search architectures respectively are significantly in excess of the $\sim$0.447 (estimated from graphs by counting pixels) reported in figure 5 of \cite{shi2020compositional} as their DLRM baseline. Our latter efficient result also uses $\sim$12$\times$ fewer parameters than the $5.4 \times 10^8$ reported for that baseline. As for search efficiency, we complete our DNAS search process in $\sim$0.28 GPU-days, which compares favorably with the $\sim$0.75 reported in \cite{song2020towards}, and especially so when we consider that they search over a dataset sub-sampled to 2M samples, which is 19.64$\times$ smaller than our search dataset (the entire training dataset). Were we to run our search over the same sub-sampled dataset, we would be 52.6$\times$ more efficient, in GPU-days, than the result reported in \cite{song2020towards}.

\section{Lessons Learned \& Insights}
\label{sec:insights}

In this section, we discuss some lessons we learned through conducting our experiments, as well as through valuable advice provided by our colleagues. We hope this might be useful to those who use our DNAS framework for their own applications, as well as other NAS researchers.

\subsection{Lessons learned}
\label{sec:insights:lessons_learned}

After experimenting with different search space groups and search spaces, we believe that the search spaces that likely demonstrate the most promise are those focused on the sparse features of the DLRM. Specifically, we believe the embedding cardinality search space may provide the best opportunities for finding excellent accuracy / model size tradeoffs, and, if properly tuned, improved, and extended, may even provide value in commercial deployments.

We also believe that much of the value of this DNAS (and tuning) framework may not be in the logloss gains, or model size reductions, but in the time it returns to industry practitioners and researchers, who can focus on model development, deployment, and so forth, instead of tuning the same model repeatedly. This improvement is impossible to quantify in a paper such as this, but we believe it is one of the key value propositions of the framework. For our own experiments, having the NAS and tuning infrastructure that we do now would have greatly accelerated our work, especially at the beginning of the project when we were less familiar with the specific architectural choices and hyperparameters that performed well for our dataset.

\subsection{Insights}
\label{sec:insights:insights}

One of the most important pieces of advice that we would like to pass on to readers is to \textbf{test search spaces (and seaarch space groups) before running NAS or designing a NAS algorithm to search over them}\footnote{Credits to Bichen Wu; see acknowledgements}. This is important because, ultimately, NAS is not magic; it can only find the best points within a search space. If the search space itself does not offer sufficient architectural variability or is not correlated with task performance or efficiency, DNAS cannot do anything to change that. For this reason, we recommend that a random search be run first, to assess the potential in any search space group.

Another piece of advice we have to offer is more of a practical recommendation when it comes to actually using DNAS, NAS, or any large-scale experimentation and tuning frameworks daily. That is to plan experiments ahead of time, review results in a timely manner, and prepare new configuration files (or however experiments are specified in a system) ahead of time. This will ensure that computational resources are fully utilized. If this is not done, it is very easy to fall behind in keeping track of experiments, resulting in unused machine time, as well as time inefficiently spent for people working on the project. This will also reduce the momentum of the project.

\section{Conclusion}
\label{sec:conclusion}

In this paper, we introduced an efficient and easily extensible Differentiable Neural Architecture Search (DNAS) framework, implemented in PyTorch and open-sourced. We described the design, structure, and functionality of this framework. This framework was then applied to one of the most commercially relevant applications of AI: ads click-through rate (CTR) prediction. Using the Deep Learning Recommendation Model (DLRM) \cite{naumov2019deep} as our backbone, we developed novel search spaces, and showed experimental results on the Criteo Kaggle dataset. These results demonstrate the promise of both our search spaces, as well as the utility of our framework. In the future, we plan to extend this work to other deep recommender backbone architectures, resulting in new search spaces. We hope this work will spur further interest in the application of NAS to CTR prediction, and that the framework we have open-sourced will allow other researchers and industry practitioners to apply the DNAS algorithm to their AI problems.

\section*{Acknowledgements}

We would like to thank Bichen Wu for helping to initiate and support this project throughout its journey, as well as for providing valuable guidance and feedback on NAS and recommendation systems (see section \ref{sec:insights:insights}). We also thank Kostadin Ilov of the ADEPT lab for help with regards to infrastructure and machines. Finally, we would like to thank Ruoxi Wang of \cite{wang2017deep} for providing helpful input regarding replication of the results from \cite{wang2017deep}.


\begin{thebibliography}{00}

\bibitem{DNASCode} \url{https://www.github.com/ravikucb/dnas}

\bibitem{wan2020fbnetv2} Wan, Alvin, Xiaoliang Dai, Peizhao Zhang, Zijian He, Yuandong Tian, Saining Xie, Bichen Wu et al. ``Fbnetv2: Differentiable neural architecture search for spatial and channel dimensions.'' In Proceedings of the IEEE/CVF Conference on Computer Vision and Pattern Recognition, pp. 12965-12974. 2020.
	
\bibitem{FBAdsAuctions} \url{https://www.facebook.com/business/help/430291176997542}

\bibitem{zhou2018deep} Zhou, Guorui, Xiaoqiang Zhu, Chenru Song, Ying Fan, Han Zhu, Xiao Ma, Yanghui Yan, Junqi Jin, Han Li, and Kun Gai. ``Deep interest network for click-through rate prediction.'' In Proceedings of the 24th ACM SIGKDD International Conference on Knowledge Discovery \& Data Mining, pp. 1059-1068. 2018.

\bibitem{guo2018visualizing} Guo, Lin, Hui Ye, Wenbo Su, Henhuan Liu, Kai Sun, and Hang Xiang. ``Visualizing and understanding deep neural networks in ctr prediction.'' arXiv preprint arXiv:1806.08541 (2018).

\bibitem{dai2020fbnetv3} Dai, Xiaoliang, Alvin Wan, Peizhao Zhang, Bichen Wu, Zijian He, Zhen Wei, Kan Chen et al. ``FBNetV3: Joint Architecture-Recipe Search using Predictor Pretraining.'' In Proceedings of the IEEE/CVF Conference on Computer Vision and Pattern Recognition, pp. 16276-16285. 2021.

\bibitem{li2015click} Li, Cheng, Yue Lu, Qiaozhu Mei, Dong Wang, and Sandeep Pandey. ``Click-through prediction for advertising in twitter timeline.'' In Proceedings of the 21th ACM SIGKDD International Conference on Knowledge Discovery and Data Mining, pp. 1959-1968. 2015.

\bibitem{wu2019efficient} Wu, Bichen. ``Efficient deep neural networks.'' arXiv preprint arXiv:1908.08926 (2019).

\bibitem{williams1992simple} Williams, Ronald J. ``Simple statistical gradient-following algorithms for connectionist reinforcement learning.'' Machine learning 8, no. 3-4 (1992): 229-256.

\bibitem{he2014practical} He, Xinran, Junfeng Pan, Ou Jin, Tianbing Xu, Bo Liu, Tao Xu, Yanxin Shi et al. ``Practical lessons from predicting clicks on ads at facebook.'' In Proceedings of the Eighth International Workshop on Data Mining for Online Advertising, pp. 1-9. 2014.

\bibitem{rong2020distributed} Rong, Haidong, Yangzihao Wang, Feihu Zhou, Junjie Zhai, Haiyang Wu, Rui Lan, Fan Li et al. ``Distributed Equivalent Substitution Training for Large-Scale Recommender Systems.'' In Proceedings of the 43rd International ACM SIGIR Conference on Research and Development in Information Retrieval, pp. 911-920. 2020.

\bibitem{smelyanskiy2019zion} Smelyanskiy, Misha. ``Zion: Facebook next-generation large memory training platform.'' In 2019 IEEE Hot Chips 31 Symposium (HCS), pp. 1-22. IEEE Computer Society, 2019.

\bibitem{he2016deep} He, Kaiming, Xiangyu Zhang, Shaoqing Ren, and Jian Sun. ``Deep residual learning for image recognition.'' In Proceedings of the IEEE conference on computer vision and pattern recognition, pp. 770-778. 2016.

\bibitem{tan2019mnasnet} Tan, Mingxing, Bo Chen, Ruoming Pang, Vijay Vasudevan, Mark Sandler, Andrew Howard, and Quoc V. Le. ``Mnasnet: Platform-aware neural architecture search for mobile.'' In Proceedings of the IEEE/CVF Conference on Computer Vision and Pattern Recognition, pp. 2820-2828. 2019.

\bibitem{devlin2019bert} Devlin, Jacob, Ming-Wei Chang, Kenton Lee, and Kristina Toutanova. ``BERT: Pre-training of Deep Bidirectional Transformers for Language Understanding.'' In Proceedings of the 2019 Conference of the North American Chapter of the Association for Computational Linguistics: Human Language Technologies, Volume 1 (Long and Short Papers), pp. 4171-4186. 2019.

\bibitem{he2019streaming} He, Yanzhang, Tara N. Sainath, Rohit Prabhavalkar, Ian McGraw, Raziel Alvarez, Ding Zhao, David Rybach et al. ``Streaming end-to-end speech recognition for mobile devices.'' In ICASSP 2019-2019 IEEE International Conference on Acoustics, Speech and Signal Processing (ICASSP), pp. 6381-6385. IEEE, 2019.

\bibitem{wu2018mixed} Wu, Bichen, Yanghan Wang, Peizhao Zhang, Yuandong Tian, Peter Vajda, and Kurt Keutzer. ``Mixed precision quantization of convnets via differentiable neural architecture search.'' arXiv preprint arXiv:1812.00090 (2018).

\bibitem{zoph2016neural} Zoph, Barret, and Quoc V. Le. ``Neural architecture search with reinforcement learning.'' arXiv preprint arXiv:1611.01578 (2016).

\bibitem{cai2018proxylessnas} Cai, Han, Ligeng Zhu, and Song Han. ``Proxylessnas: Direct neural architecture search on target task and hardware.'' arXiv preprint arXiv:1812.00332 (2018).

\bibitem{naumov2019deep} Naumov, Maxim, Dheevatsa Mudigere, Hao-Jun Michael Shi, Jianyu Huang, Narayanan Sundaraman, Jongsoo Park, Xiaodong Wang et al. ``Deep learning recommendation model for personalization and recommendation systems.'' arXiv preprint arXiv:1906.00091 (2019).

\bibitem{zhou2019deep} Zhou, Guorui, Na Mou, Ying Fan, Qi Pi, Weijie Bian, Chang Zhou, Xiaoqiang Zhu, and Kun Gai. ``Deep interest evolution network for click-through rate prediction.'' In Proceedings of the AAAI conference on artificial intelligence, vol. 33, no. 01, pp. 5941-5948. 2019.

\bibitem{wang2017deep} Wang, Ruoxi, Bin Fu, Gang Fu, and Mingliang Wang. ``Deep \& cross network for ad click predictions.'' In Proceedings of the ADKDD'17, pp. 1-7. 2017.

\bibitem{pi2019practice} Pi, Qi, Weijie Bian, Guorui Zhou, Xiaoqiang Zhu, and Kun Gai. ``Practice on long sequential user behavior modeling for click-through rate prediction.'' In Proceedings of the 25th ACM SIGKDD International Conference on Knowledge Discovery \& Data Mining, pp. 2671-2679. 2019.

\bibitem{wu2019fbnet} Wu, Bichen, Xiaoliang Dai, Peizhao Zhang, Yanghan Wang, Fei Sun, Yiming Wu, Yuandong Tian, Peter Vajda, Yangqing Jia, and Kurt Keutzer. ``Fbnet: Hardware-aware efficient convnet design via differentiable neural architecture search.'' In Proceedings of the IEEE/CVF Conference on Computer Vision and Pattern Recognition, pp. 10734-10742. 2019.

\bibitem{ishkhanov2020time} Ishkhanov, Tigran, Maxim Naumov, Xianjie Chen, Yan Zhu, Yuan Zhong, Alisson Gusatti Azzolini, Chonglin Sun, Frank Jiang, Andrey Malevich, and Liang Xiong. ``Time-based Sequence Model for Personalization and Recommendation Systems.'' arXiv preprint arXiv:2008.11922 (2020).

\bibitem{shi2020compositional} Shi, Hao-Jun Michael, Dheevatsa Mudigere, Maxim Naumov, and Jiyan Yang. ``Compositional embeddings using complementary partitions for memory-efficient recommendation systems.'' In Proceedings of the 26th ACM SIGKDD International Conference on Knowledge Discovery \& Data Mining, pp. 165-175. 2020.

\bibitem{song2020towards} Song, Qingquan, Dehua Cheng, Hanning Zhou, Jiyan Yang, Yuandong Tian, and Xia Hu. ``Towards automated neural interaction discovery for click-through rate prediction.'' In Proceedings of the 26th ACM SIGKDD International Conference on Knowledge Discovery \& Data Mining, pp. 945-955. 2020.

\bibitem{song2019autoint} Song, Weiping, Chence Shi, Zhiping Xiao, Zhijian Duan, Yewen Xu, Ming Zhang, and Jian Tang. ``Autoint: Automatic feature interaction learning via self-attentive neural networks.'' In Proceedings of the 28th ACM International Conference on Information and Knowledge Management, pp. 1161-1170. 2019.

\bibitem{gupta2020deeprecsys} Gupta, Udit, Samuel Hsia, Vikram Saraph, Xiaodong Wang, Brandon Reagen, Gu-Yeon Wei, Hsien-Hsin S. Lee, David Brooks, and Carole-Jean Wu. ``Deeprecsys: A system for optimizing end-to-end at-scale neural recommendation inference.'' In 2020 ACM/IEEE 47th Annual International Symposium on Computer Architecture (ISCA), pp. 982-995. IEEE, 2020.

\bibitem{gupta2020architectural} Gupta, Udit, Carole-Jean Wu, Xiaodong Wang, Maxim Naumov, Brandon Reagen, David Brooks, Bradford Cottel et al. ``The architectural implications of facebook's dnn-based personalized recommendation.'' In 2020 IEEE International Symposium on High Performance Computer Architecture (HPCA), pp. 488-501. IEEE, 2020.

\bibitem{he2017neural} He, Xiangnan, Lizi Liao, Hanwang Zhang, Liqiang Nie, Xia Hu, and Tat-Seng Chua. ``Neural collaborative filtering.'' In Proceedings of the 26th international conference on world wide web, pp. 173-182. 2017.

\bibitem{goldberg1992using} Goldberg, David, David Nichols, Brian M. Oki, and Douglas Terry. ``Using collaborative filtering to weave an information tapestry.'' Communications of the ACM 35, no. 12 (1992): 61-70.

\bibitem{rendle2010factorization} Rendle, Steffen. "Factorization machines." In 2010 IEEE International Conference on Data Mining, pp. 995-1000. IEEE, 2010.

\bibitem{chen2020adabert} Chen, Daoyuan, Yaliang Li, Minghui Qiu, Zhen Wang, Bofang Li, Bolin Ding, Hongbo Deng, Jun Huang, Wei Lin, and Jingren Zhou. ``Adabert: Task-adaptive bert compression with differentiable neural architecture search.'' arXiv preprint arXiv:2001.04246 (2020).

\end{thebibliography}
\end{document}